\documentclass{egpubl}
\usepackage{egsgp2022}
\usepackage[T1]{fontenc}
\usepackage{dfadobe}  

\BibtexOrBiblatex
\electronicVersion
\PrintedOrElectronic

\usepackage{egweblnk} 
\usepackage{my}

\title[SDF-StyleGAN: Implicit SDF-Based StyleGAN for 3D Shape Generation]{SDF-StyleGAN: Implicit SDF-Based StyleGAN \\for 3D Shape Generation}

\author[Xin-Yang Zheng \& Yang Liu \& Peng-Shuai Wang \& Xin Tong]
{
\parbox{\textwidth}{\centering Xin-Yang Zheng\thanks{Work done during an internship at Microsoft Research Asia}$^{1}$\orcid{0000-0003-2318-1863}, Yang Liu\thanks{Corresponding author}$^2$ \orcid{0000-0002-3768-6654}, Peng-Shuai Wang$^2$ \orcid{0000-0001-9700-8188}, and Xin Tong$^2$\orcid{0000-0001-8788-2453}}
        \\
{\parbox{\textwidth}{\centering $^1$ Tsinghua University, P.~R.~China, \qquad
         $^2$ Microsoft Research Asia}}
}

\begin{document}

\teaser{
    \includegraphics[width=1\linewidth]{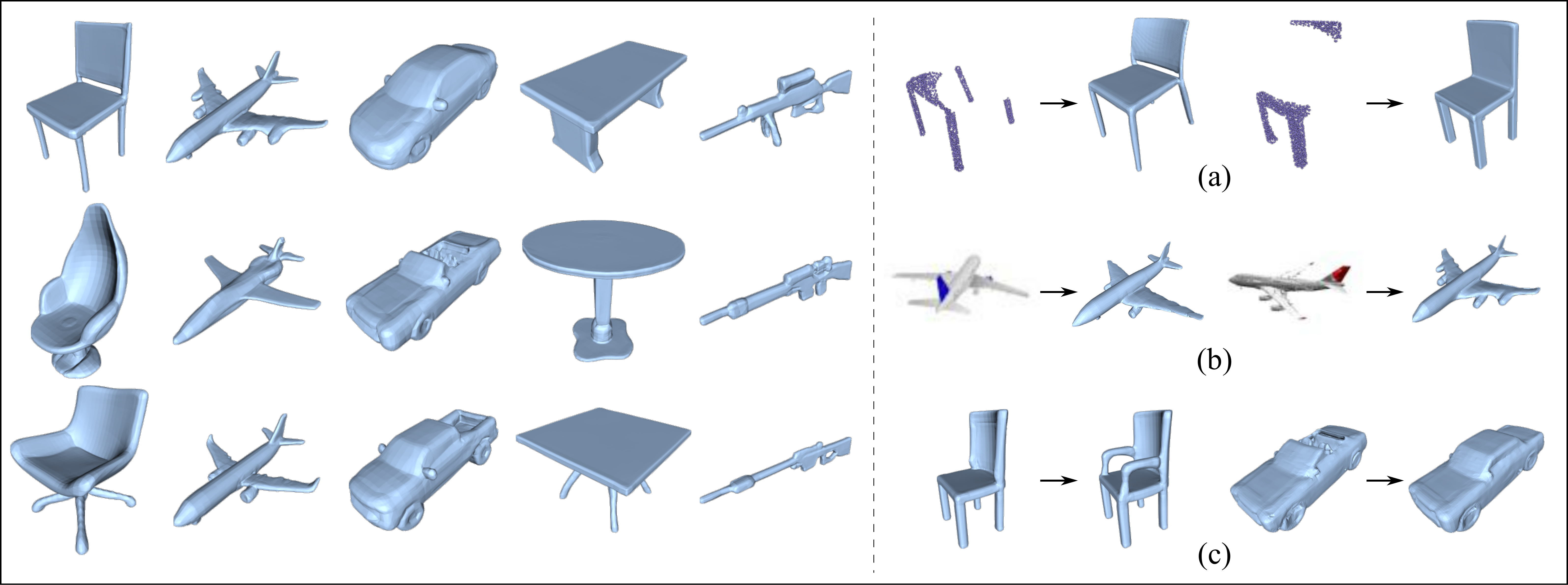}
    \centering
    \caption{\textbf{Left}: diverse and visually plausible 3D shapes generated by SDF-StyleGAN. \textbf{Right}: the applications enabled by SDF-StyleGAN. (a) Shape completion from partial point cloud inputs. (b) Shape generation from single-view images. (c) Chair arms and car roofs are generated via shape style editing.}
    \label{fig:teaser}
}

\maketitle
\begin{abstract}
We present a StyleGAN2-based deep learning approach for 3D shape generation, called \emph{SDF-StyleGAN}, with the aim of reducing visual and geometric dissimilarity between generated shapes and a shape collection. We extend StyleGAN2 to 3D generation and utilize the implicit signed distance function (SDF) as the 3D shape representation, and introduce two novel global and local shape discriminators that distinguish real and fake SDF values and gradients to significantly improve shape geometry and visual quality. We further complement the evaluation metrics of 3D generative models with the shading-image-based Fr\'echet inception distance (FID) scores to better assess visual quality and shape distribution of the generated shapes. Experiments on shape generation demonstrate the superior performance of SDF-StyleGAN over the state-of-the-art. We further demonstrate the efficacy of SDF-StyleGAN in various tasks based on GAN inversion, including shape reconstruction, shape completion from partial point clouds, single-view image-based shape generation, and shape style editing. Extensive ablation studies justify the efficacy of our framework design. Our code and trained models are available at  \url{https://github.com/Zhengxinyang/SDF-StyleGAN}.


\end{abstract}

\section{Introduction} \label{sec:intro}

Automatic and controllable generation of high-quality 3D shapes is one of the key tasks of computer graphics, as it is useful to enrich 3D model collections, simplify the 3D modeling and reconstruction process, and to provide a large amount of diverse 3D training data to improve shape recognition and other 3D vision tasks. Inspired by the success of generative models such as Variational AutoEncoder (VAE)~\cite{kingma2013auto}, Generative Adversarial Networks (GAN)~\cite{goodfellow2014generative} in image generation and natural language generation, the 3D shape generation paradigm has been shifted to learning-based generative models in recent years.

Previous learning-based 3D generation works can be differentiated by the chosen 3D shape representations and generative models. Point cloud representation is popular in 3D generation. However, due to its discrete nature, the generative shapes do not have continuous 3D geometry, and their conversion to 3D surfaces is also non-trivial. Mesh representation overcomes this limitation by modeling 3D shapes as a piecewise-linear surface, but the modeling capability is usually restricted by the predefined mesh template and cannot allow the change of mesh topology easily. Volumetric occupancy or signed distance function treats 3D shapes as the isosurfaces of an implicit function, and the recent development of neural implicit representation such as DeepSDF~\cite{park2019deepsdf} offers a convenient shape representation for 3D generation. Popular generative models such as VAE, GAN, flow-based generative models~\cite{rezende2015variational}, and autoregressive models~\cite{van2016pixel} have been incorporated with different 3D representations.

Despite the rapid development of 3D generative models, we observed that there are clear visual and geometric differences, between generative shapes and real shape collection in existing approaches. For instance, shapes generated by many existing works often exhibit bumpy and incomplete geometry, which can be found in \cref{fig:chaircomp} and \cref{fig:otherclass}. In contrast to 3D generation, the gap between real and generative data in the image generation domain becomes much smaller by using more advanced generative models such as StyleGAN~\cite{karras2019style} and its successors~\cite{karras2020stylegan2,karras2021alias}. We attribute this performance gap in 3D generation to the following three aspects. First, the design of current 3D generators is relatively weaker than the state-of-the-art image generation networks; secondly, the works based on point cloud and mesh representation have limited shape representation power; thirdly, the assessment of generated shape quality that defines the discrimination loss of GAN is not sensitive to visual appearance, which is important to human perception. To this end, in the presented work, we use the StyleGAN2 network~\cite{karras2020stylegan2} to enhance the 3D shape generator, employ grid-based implicit SDF to improve shape modeling ability, and propose novel SDF discriminators that assess the SDF values and its gradient values from global and local views to improve visual and geometric similarity between the generated shapes and the training data. In particular, the use of SDF gradients greatly enhances surface quality and visual appearance, as SDF gradients determine surface normals that contribute to visual perception via rendering.

We validate the design choice of our 3D generative network --- called \emph{SDF-StyleGAN}, through extensive ablation studies, and demonstrate the superiority of \emph{SDF-StyleGAN} to other state-of-the-art 3D shape generative works via qualitative and quantitative comparisons. To measure visual quality and shape distribution of the generated shapes, we propose shading-image-based Fr\'echet inception distance (FID) scores that complement the evaluation metrics for 3D generative models. We also present various applications enabled by \emph{SDF-StyleGAN}, including shape reconstruction, shape completion from point clouds, single-view image-based shape generation, and shape style editing.
\section{Related Work} \label{sec:related}
\subsection{3D generative models}\label{subsec:modelreview}

\myparagraph{GAN-based models}
GAN models were first introduced to voxel-based 3D generation by Wu \etal~\cite{wu2016learning} and their training was improved by adapting the Wasserstein distance~\cite{smith2017improved,huang20193d}. Achlioptas \etal~\cite{achlioptas2018learning} brought GAN to point cloud generation and also invented l-GAN which first trains an autoencoder (AE) and then trains a GAN on the latent space of AE. Jiang and Marcus~\cite{jiang2017hierarchical} used both the low-frequency generator and the high-frequency generator to improve the quality of the generated discrete SDF grids. IMGAN~\cite{chen2019learning} integrated l-GAN and neural implicit representation to achieve better performance.
Kleineberg \etal~\cite{kleineberg2020adversarial} experimented with both implicit SDF and point cloud as shape representations in a GAN model, which uses a 3DCNN-based discriminator or a PointNet~\cite{qi2017pointnet}-based discriminator. Hui \etal~\cite{hui2020progressive} proposed to generate multiresolution point clouds progressively and use adversarial losses in multiresolution to improve point cloud quality. Wen \etal~\cite{wen2021learning} also generated points progressively via a dual generator framework.  Ibing \etal~\cite{ibing20213d} localized l-GAN on grid cells and used the implicit occupancy field as its shape representation. Li \etal~\cite{li2019synthesizing}, Wu \etal~\cite{wu2020dfr} and Luo \etal~\cite{luo2021surfgen} used 2D differentiable rendering for training 3D GANs without any 3D supervision. The former two works used silhouette images, while Luo \etal~used a designed spherical map of the extracted mesh. Shu \etal~\cite{mo2020pt2pc} introduced a tree-structured graph convolution network as the generator to improve feature learning. For generating part-controllable shapes, Wang \etal~\cite{wang2018global} introduced a global-to-local GAN in which the generator also produces segmentation labels and each local discriminator is designed for each segmented part.  Shape part information was also utilized to train the network \cite{dubrovina2019composite,li2020learning} to generate semantically plausible shapes. To remove the requirement of part annotation, the StyleGAN-like architecture was adapted for part-level controllable point cloud generation~\cite{gal2021mrgan,li2021sp}. For better controlling the generated 3D model, Chen \etal~\cite{chen2021decor} proposed a GAN model to refine a low-resolution coarse voxel shape into a high-resolution model with finer details. With the use of volume rendering and neural implicit 3D representations, GAN models are also used to synthesize 3D-aware images such as human faces~\cite{or2021stylesdf,chan2021pi} and achieve high quality results.  In our work, we also adapt the StyleGAN architecture to generate feature vectors on grids that are mapped to implicit SDFs, and use SDF gradients to guide the global and local discriminators to improve shape quality and visual appearance.

\myparagraph{Autoencoder-based models}
AtlasNet~\cite{groueix2018papier} encoded 3D shapes into a latent space and took a collection of parametric surface elements as shape representations for shape generation. Li \etal~\cite{li2017grass} created a recursive network based on variational autoencoder (VAE) to generate 3D shapes with hierarchical tree structures. Mo \etal~\cite{mo2019structurenet} encoded more structural relations with hierarchical graphs and devised a Graph-CNN-based VAE for shape generation. Gao \etal~\cite{gao2019sdm} proposed a two-level VAE to learn the geometry of the deformed parts and the shape structure. Jie \etal~\cite{Jie20DsgNet} combined the advantages of the above two approaches to learn the disentangled shape structure and geometry.  Wu \etal~\cite{wu2020pq} used a Seq2Seq Autoencoder to generate 3D shapes via sequential part assembly. In these works, part information is needed for training.  Li \etal~\cite{li2021editvae} introduced superquadric primitives in VAE learning to mimic shape parts for easy editing, without real part annotation.

\myparagraph{Other generative models}
By transforming a complex distribution to a simple and predefined distribution, diffusion-based generative models~\cite{song2019generative, ho2020denoising} have been applied to point cloud generation via learning the reverse diffusion process that maps a noise distribution to a 3D point cloud~\cite{cai2020learning,luo2021diffusion,zhou20213d}. By explicitly modeling a probability distribution via normalizing flow, flow-based generative models~\cite{rezende2015variational} are also popular in shape generation, especially point cloud generation~\cite{yang2019pointflow,klokov2020discrete,kim2020softflow,pumarola2020c,kimura2021chartpointflow,postels2021go}. However, these works focus on point clouds and do not generate continuous shapes directly. By implicitly defining a distribution over sequences using the chain rule for conditional probability, autoregressive models achieve their success in image generation~\cite{van2016pixel}. They were also used in point cloud generation~\cite{sun2018pointgrow}, octree generation~\cite{ibing2021octree}, polygonal mesh generation~\cite{nash2020polygen}, and SDF generation~\cite{autosdf2022}.   By explicitly modeling a probability distribution in the form of the EBM~\cite{xie2016theory}, Xie \etal~\cite{xie2020generative,xie2021generative} used an energy-based generative model to synthesize 3D shapes via Markov chain Monte Carlo sampling. \looseness=-1

\begin{figure*}[t]
    \centering
    \includegraphics[width=1\linewidth]{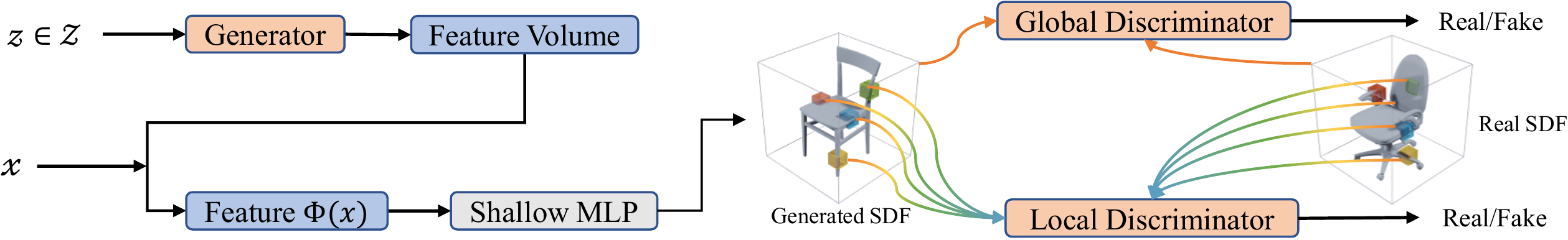}
    \caption{Overview of SDF-StyleGAN. The original StyleGAN2 generator is extended to 3D, and it outputs the feature volume in a unit box. The feature vector $\phi(\mx)$ at any point $\mx$ inside the volume is interpolated via trilinear interpolation and is mapped to SDF value via a shallow MLP. The global discriminator takes the SDF values and gradients sampled at the grid centers as input, and the local discriminator takes the SDF values and gradients at a local and random 3D box region near the surface as input. A few local box regions are illustrated on the above chair example, in different colors.}
    \label{fig:sdfstylegan} \vspace{-5mm}
\end{figure*}

\subsection{Evaluation metrics for 3D generation} \label{subsec:metricreview}

To measure the similarity between the generated point cloud data set $S_g$ and the referenced point cloud data set $S_r$, Achlioptas \etal~\cite{achlioptas2018learning} proposed three metrics: Coverage (COV) that is the fraction of $S_r$ covered by $S_g$; Minimum Matching Distance (MMD) that measures how well the covered shapes in $S_r$ are represented by $S_g$; and Jensen-Shannon Divergence (JSD) between two probability distributions over $S_g$ and $S_r$. The distance between two point clouds can be either the Chamfer distance (CD) or the Earth Mover's distance (EMD). Chen \etal~\cite{chen2019learning} replaced CD and EMD with light-field-descriptor (LFD)~\cite{chen2003visual} for better measuring the quality of the generated meshes.
Noticing that COV, MMD and JSD do not ensure a fair model comparison,  Yang \etal~\cite{yang2019pointflow} proposed  1-Nearest Neighbor Accuracy (1-NNA) to assess whether two distributions are identical. Ibing \etal~\cite{ibing20213d} also adapted a robust statistical criterion --- Edge Count Difference (ECD)~\cite{chen2017new} to evaluate 3D generative models.

To mimic the Fr\'echet Inception Distance (FID)~\cite{heusel2017gans} that is widely used in image generation for assessing image quality, Shu \etal~\cite{shu20193d} proposed the  Fr\'echet Point Cloud Distance (FPD) that computes the 2-Wasserstein distance between real and fake Gaussian measures in the feature space extracted by a pretrained PointNet~\cite{qi2017pointnet}. Li \etal~\cite{li2021sp} replaced PointNet with a stronger pretrained backbone -- DGCNN~\cite{wang2019dynamic} for FPD computation.

In addition, some works propose task-specific metrics. Wang \etal~\cite{wang2018global} proposed the 3D inception score, the symmetry score, and the distribution distance to measure the quality of the generated voxelized shapes and generated parts, Mo \etal~\cite{mo2020pt2pc} proposed the HierInsSeg score to measure how well the generated point clouds satisfy the part tree conditions they defined. Human preference through user studies is also used for assessing the quality of shape generation~\cite{kim2020softflow,li2021sp}.

Other metrics based on shading images for measuring shape reconstruction quality, such as the mean square error defined over per-pixel keypoint maps~\cite{jin2020dr}, have potentials to be utilized for evaluating 3D generation.  In our work, we enrich the evaluation metrics by introducing Fr\'echet Inception Distance on shading images, to better assess the visual appearance and data distribution of 3D generative models.

\section{Design of SDF-StyleGAN} \label{sec:method}

\subsection{Overview}
We design a StyleGAN2-based 3D shape generation architecture, as illustrated in \cref{fig:sdfstylegan}. We use feature-volume-based implicit signed distance functions as shape representation to maximize implicit representation capability (\cref{subsec:featurevolume}), leverage and extend the 2D StyleGAN2 generator~\cite{karras2020stylegan2} to 3D to generate 3D feature volumes (\cref{subsec:generator}), and propose global and local discriminators to distinguish \emph{fake} SDF fields, including their gradients calculated from global and local regions of the generated field, from the sampled SDF field of the \emph{real} shapes in the training set (\cref{subsec:discriminator}). We propose an adaptive training scheme to gradually improve the quality of the generated shapes (\cref{subsec:training}).

\subsection{Feature-volume-based implicit signed distance functions} \label{subsec:featurevolume}

We represent 3D shapes as feature-volume-based implicit signed distance functions, in similar spirit to the grid-based implicit occupancy function used in~\cite{jiang2020local,peng2020convolutional,ibing20213d}.

We assume that a 3D shape $\mS$ is generated inside a box $B: [-1,1]^3$ and the box is divided as $m \times m \times m $ grid cells with equal cell size. \begin{wrapfigure}[9]{r}{0.45\columnwidth}
  \begin{center}
    \begin{overpic}[width=0.35\columnwidth]{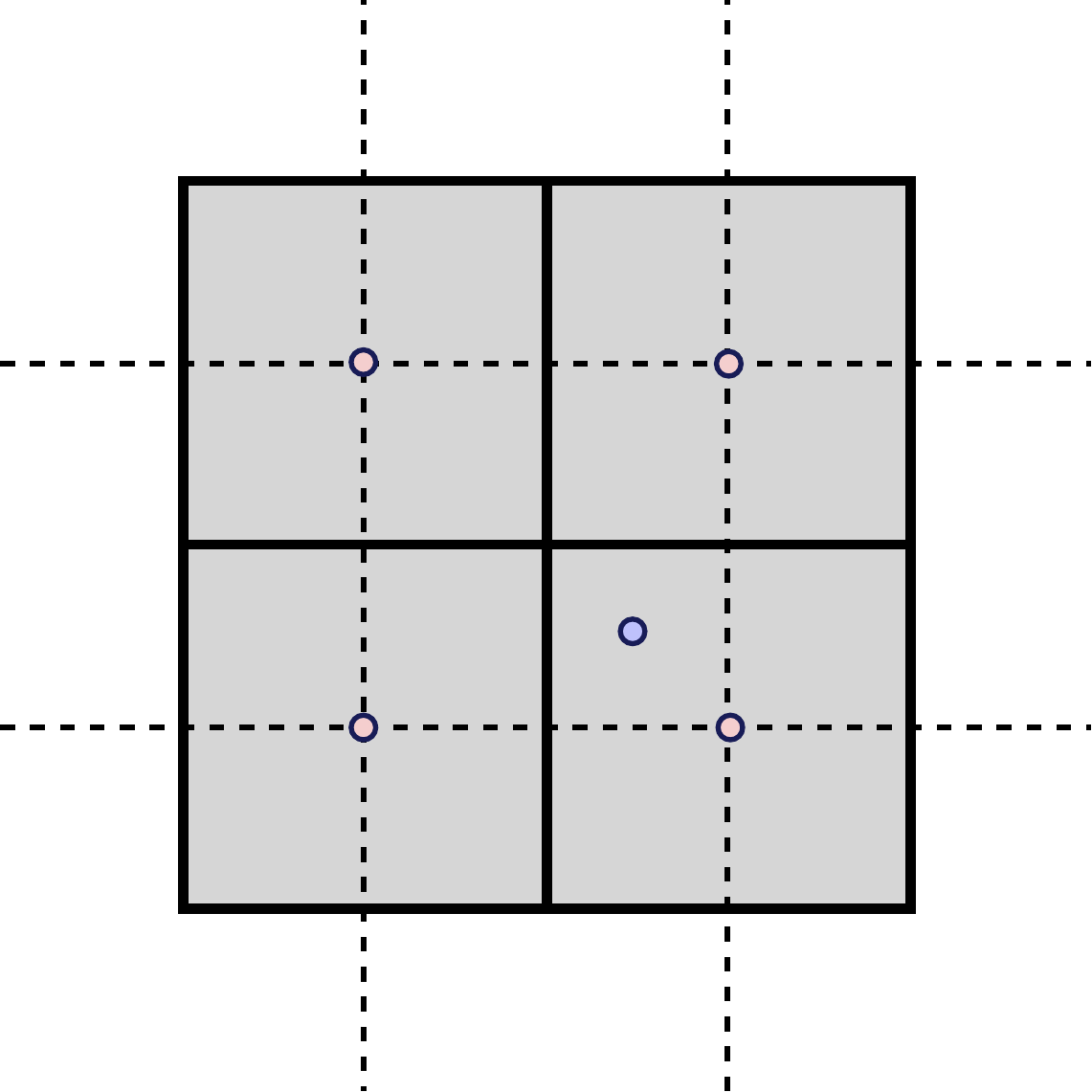}
      \put(53,44){\scriptsize $\Phi(\mx)$}
      \put(24,70){\scriptsize $\Phi_1$}
      \put(68,70){\scriptsize $\Phi_2$}
      \put(24,27){\scriptsize $\Phi_3$}
      \put(68,27){\scriptsize $\Phi_4$}
    \end{overpic}
  \end{center}
\end{wrapfigure}

A dual graph is constructed on these grid cells, and on each graph node $c_i$, a feature vector $\Phi_i \in \mathbb{R}^l$ is stored. All feature vectors form a feature volume.
For any 3D point $\mx \in B$, we assign a feature vector to it, denoted as $\Phi(\mx)$, via trilinear interpolation in the feature volume, \ie using the feature vectors of the eight corners of the dual cell where $\mx$ is located. Due to the construction above, $\Phi(\mx)$ is a continuous function defined in $B$.
A shallow multilayer perceptron (MLP) is used to map $\Phi(\mx)$ to a signed distance value to the shape. In this way, an SDF $f(\mx): \mx \in \mathbb{R}^3 \longmapsto \mathbb{R}$ is determined by the feature volume and the MLP mapping.
The zero isosurface of the SDF defines the shape geometry. In the right inset, we illustrate the concept of feature volume in 2D. The gray grid is a portion of the primal grid cells and the dashed black grid is its dual graph. To calculate the feature at point $\mx$, we first find the dual cell where $\mx$ is located and use the feature vectors $\Phi_1,\Phi_2,\Phi_3,\Phi_4$ at the dual cell vertices to
perform the interpolation.

In our implementation, $m=32$, $l=16$ and the MLP has two hidden layers. The feature volume is produced by the generator, and the MLP parameters are learned via training.

\subsection{SDF-StyleGAN generator} \label{subsec:generator}

\begin{figure*}[t]
    \centering
    \includegraphics[width=0.95\linewidth]{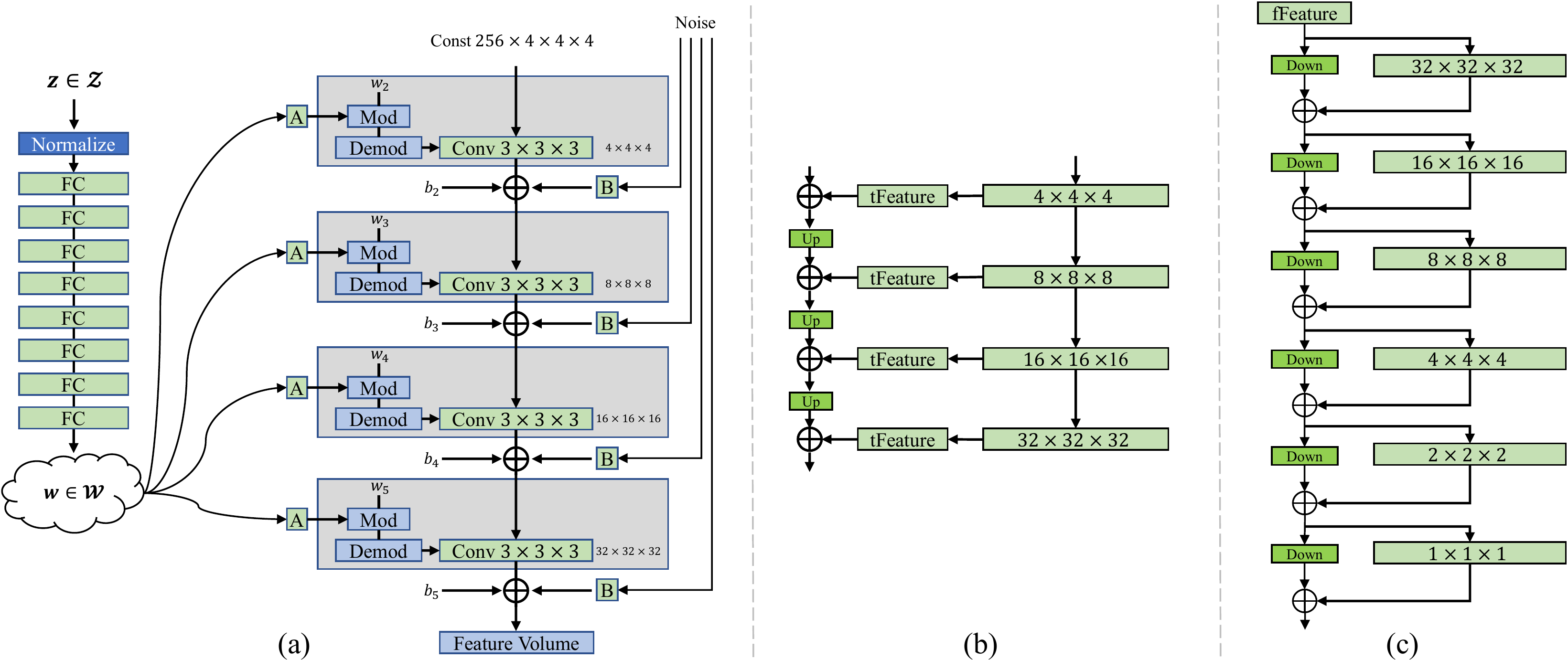}
    \caption{(a): The revised StyleGAN2 generator for 3D feature volume generation. We used 3D convolution with kernel size 3 and four style blocks corresponding to four-level resolution, up to $32\times32\times 32$.  \emph{Mod} and \emph{Demod} are the modulation and demodulation modules adapted from StyleGAN2. (b): The skip input for the generator. (c) The discriminator architecture. The \emph{tFeature} module and the \emph{fFeature} module convert between the feature volume per grid cell and the high dimensional feature to/from 3D convolution. \emph{Up} and \emph{Down} denote the upsampling and downsampling modules. The first block in (c) is removed from the local discriminator as its input feature grid resolution is $16\times16\times16$.}
    \label{fig:stylegandetail} \vspace{-5mm}
\end{figure*}

We adapt the StyleGAN2 architecture~\cite{karras2020stylegan2} to our feature volume generator. In the original StyleGAN2 architecture, a latent vector $\mz$ is first sampled from a high-dimensional normal distribution and mapped to an intermediate latent space $\mW$. A learned affine transform and Gaussian noise are injected into a style block, and multiple stacked style locks are used to predict the pixel colors. We applied the following modifications to the original StyleGAN2's network structure.
\begin{inparaenum}[(i).]
    \item The 2D convolution in each StyleGAN2 block is changed to 3D convolution with corresponding kernel size;
    \item The stacked blocks are upsampled from $4\times4\times4$ to $m\times m\times m$;
    \item the size of the constant tensor fed to the first style block is set to $256\times4\times4\times4$.
\end{inparaenum}
The detailed structure of the network and our changes are illustrated in \cref{fig:stylegandetail}-(a,b).  With these modifications, the generator output an $m \times m \times m$ feature volume with $l$ channels.

\subsection{SDF-StyleGAN discriminator} \label{subsec:discriminator}
We design two discriminators to evaluate the generated SDF on the coarse scale and the fine scale: global discriminator and local discriminator. Noticing that the visual appearance of the zero isosurface is affected by the surface normal, we also propose to take both SDF values and SDF gradients on a grid region as input to discriminators because the SDF gradients at the zero isosurface are exactly shape surface normals, and the use of both zero-order and first-order SDF information on a coarse grid can approximate SDF values on a much denser grid.

\subsubsection{Global discriminator} For a generated feature volume, we calculate the SDF values and gradients in a coarse grid of $[-1,1]^3$ with resolution $K_G \times K_G \times K_G $. The SDF values are evaluated directly via the network and SDF gradients are calculated via finite differentiation. These samples form a $4 \times K_G \times K_G \times K_G $ feature grid that can approximate the shape roughly. For a \emph{real} shape in the training set, we first build a discrete SDF field with resolution $128 \times 128 \times 128$, then sample the SDF values on the same coarse grid via trilinear interpolation. The corresponding SDF gradients are approximated via finite differentiation. As SDF gradient vectors should be with unit length in theory, we normalize the gradients before constructing the feature grid. In our implementation, we set $K_G = m = 32$.

We adapt the discriminator architecture of StyleGAN2 to our global discriminator $\mathcal{D}_G$ by replacing the 2D convolution with the corresponding 3D convolution, as illustrated in \cref{fig:stylegandetail}-(c). The discriminator is used to distinguish between the synthesized shapes and the shapes from the training set.

\subsubsection{Local discriminator}
The global discriminator focuses on shape quality in a coarse level. Naively increasing $K_G$ will lead to high computational and memory costs due to the use of 3D convolution, and the discrimination of \emph{real} or \emph{fake} SDF away from the zero isosurface has little impact on the quality of shape surface. Instead, we pay more attention to the local region close to the zero isosurface by introducing a local discriminator $\mathcal{D}_L$.

We choose a set of local regions as follows. We first evaluate the SDF values on the cell centers of the coarse grid used in the global discriminator and sort these cell centers in ascending order according to their absolute SDF values. The first $N_0$ cell centers are selected as the centers of candidate local regions. We then randomly select $s$ cell centers from these candidates and construct a small box with length $L_b$ centered at every selected center. For each small box, we divide it as a $K_L \times K_L \times K_L$ grid, sample the SDF values and calculate gradients at the cell centers of this grid to construct a $4 \times K_L \times K_L \times K_L$ feature grid.  The local discriminator takes the feature grid as input, to distinguish whether it is \emph{real} or \emph{fake}.

Note that we did not choose the first $s$ cell centers from the sorted center list, as this greedy selection may result in clustered local regions. Our random selection from a large candidate pool helps to distribute local regions more evenly around the zero isosurface. In our implementation, the indices of $N_0$ candidates are randomized directly for selection. It is possible to improve this randomization by using the furthest point sampling strategy, but it requires more computational time for training.

The network architecture of the local discriminator is similar to that of the global discriminator. During training, each \emph{real} or \emph{generated} shape provides $s$ local regions to the local discriminator.
In our implementation, we set $K_L=16$, $L_b = 0.25$, $N_0 = 2048$ and $s = 16$ by default. In \cref{fig:sdfstylegan}, we illustrate some local region boxes in the chair examples.

\subsection{SDF-StyleGAN training} \label{subsec:training}

\subsubsection{Loss functions}
The loss functions of our discriminators and generator are adapted from StyleGAN2.
The global discriminator loss $\mathcal{L}^G_{\mathcal{D}}$ has the following form:
\begin{equation}
    \begin{aligned}
        \mathcal{L}^G_{\mathcal{D}} & = \mathbb{E}_{\mz\sim \mathcal{Z}}[\zeta(\mathcal{D}_G(\mat{T}_G(\mathcal{S}(\mz))))] +\mathbb{E}_{\mathcal{S}\sim p_{data}}[\zeta(-\mathcal{D}_G(\mat{T}_G(\mathcal{S})))] \\ &+ \mathcal{R}_{\mathcal{S}\sim p_{data}}(\mathcal{S};\Theta_G).
    \end{aligned}
    \label{eq:global_discriminator_loss}
\end{equation}
Here, $\mathcal{S}(\mz)$ and $\mathcal{S}$ denote the generated SDF and the SDF of the shape sampled from the training data set respectively. $\mathcal{D}_G(\cdot)$ is the output of the global discriminator, and $\mat{T}_G(\cdot)$ denotes the feature grid (SDF values \& gradients) calculated in the coarse grid.
$\zeta(x) = \log(1 + e^x)$.  $\mathcal{R}_{S\sim p_{data}}(S;\Theta)$ is the $R_1$ regularization term adapted from \cite{mescheder2018training}, where $\Theta_G$ is the network parameters of $\mathcal{D}_G$.

The local discriminator loss $\mathcal{L}^L_{\mathcal{D}}$ is similar to $\mathcal{L}^G_{\mathcal{D}}$.
\begin{equation}
    \begin{aligned}
        \mathcal{L}^L_{\mathcal{D}} & = \mathbb{E}_{\mz\sim \mathcal{Z}}[\zeta(\mathcal{D}_L(\mat{T}_L(\mathcal{S}(\mz))))] +\mathbb{E}_{\mathcal{S}\sim p_{data}}[\zeta(-\mathcal{D}_L(\mat{T}_L(\mathcal{S})))] \\ &+ \mathcal{R}_{S\sim p_{data}}(\mathcal{S};\Theta_L).
    \end{aligned}
    \label{eq:local_discriminator_loss}
\end{equation}
Here $\mat{T}_L(\cdot)$ is the feature grid calculated in the local region, and $\Theta_L$ is the network parameters of the local discriminator. \looseness=-1

The loss function of the generator $\mathcal{L}_{\mathcal{G}}$  is defined as follows.
\begin{equation}
    \begin{aligned}
        \mathcal{L}_{\mathcal{G}} & =  \beta \, \mathcal{L}_{\mathrm{pl}} +
        \mathbb{E}_{\mz\sim \mathcal{Z}}[\zeta(-\mathcal{D}_G(\mat{T}_G(\mathcal{S}(\mz))))] \\ & +
        \alpha(t) \, \mathbb{E}_{\mz\sim \mathcal{Z}}[\zeta(-\mathcal{D}_L(\mat{T}_L(\mathcal{S}(\mz))))].
    \end{aligned}
    \label{eq:generator_loss}
\end{equation}
Here, $\mathcal{L}_{\mathrm{pl}}$ is the path length regularization term used by StyleGAN2, and $\beta = 2$. $\alpha(t)$ is the weight of the local discriminator and changed during training. We also adopt the EMA scheme~\cite{yaz2018unusual} to stabilize the generator training.

\subsubsection{Adaptive training scheme}
We propose the following adaptive scheme to train SDF-StyleGAN. We first initialize the generator and discriminators with StyleGAN2's weight initialization, then update the global discriminator, the local discriminator, and the generator in sequential order. During the early training stage, we set a small $\alpha$ in generator training so that the global discriminator dominates, and the network focuses on generating rough SDF fields. We gradually increase $\alpha$ so that the local discriminator can improve the geometry details while maintaining the global shape structure. In our implementation, we set $\alpha(t) = \alpha_{\min} + \alpha_{\max} \times t/t_{\max}$, here $t$ is the current epoch number and $t_{\max}$ is the maximum epoch number. Our default setting is $\alpha_{\min}=0, \alpha_{\max}=8, t_{\max}=200$.

\section{Experiments and Evaluation} \label{sec:eval}

\myparagraph{Dataset and training}
We trained our SDF-StyleGAN with the five shape categories selected from ShapeNet Core V1~\cite{chang2015shapenet}: chair, table, airplane, car, and rifle, individually. We use the same data split of ~\cite{chen2019learning}: \SI{70}{\percent} data as the training set, \SI{20}{\percent} data as the test set, and \SI{10}{\percent} data as the validation set which is not used in our approach. For each shape with triangle mesh format in the training set, we normalize it into a $[-0.8,0.8]^3$ box, and use the SDF computation algorithm of ~\cite{xu2014signed} to compute the discrete SDF field with resolution $128^3$ in $[-1,1]^3$. This algorithm can remove nested interior mesh facets, handle non-watertight meshes and meshes with inconsistently oriented normals robustly. During training, we also ensure that the centers of the selected local regions are contained in $[-1+L_b/2,1-L_b/2]^3$, so that all the selected local regions are strictly within $[-1,1]^3$.   We conducted our experiments on a Linux server with Intel Xeon Platinum 8168 CPU (\SI{2.7}{GHz}) and 8 Tesla V100 GPUs (\SI{16}{GB} memory). The default batch size is 32. The training time (200 epochs) takes about 2 days on average. \cref{fig:teaser} illustrates a few plausible shapes generated by our approach.

\myparagraph{Competitive methods} We select the following representative works that generate continuous 3D shapes for comparison: IMGAN~\cite{chen2019learning} and Implicit-Grid~\cite{ibing20213d} that use implicit occupancy fields to represent 3D shapes, ShapeGAN~\cite{kleineberg2020adversarial} that uses implicit SDFs to represent 3D shapes. We reused the released checkpoints of IMGAN and Implicit-Grid for evaluation. More specifically, IMGAN~\cite{chen2019learning}'s chair and table categories were trained on $64^3$ resolution, while airplane, car and rifle categories were trained on $128^3$ resolution. For Implicit Grid~\cite{ibing20213d}, we used the checkpoints provided by the authors, trained on $256^3$ resolution. As the work of \cite{kleineberg2020adversarial} released the pre-trained networks in three ShapeNetCore V2 shape categories only, for a fair comparison, we retrain ShapeGAN on our processed data with the same training strategy and parameters provided by the authors.

In the following subsections, we first present the evaluation metrics for 3D shape generation in \cref{subsec:metric}, then provide a quantitative and qualitative evaluation of our method and the competitive methods in \cref{subsec:comp}. Ablation studies on the design of SDF-StyleGAN are presented in \cref{subsec:ablation}.

\subsection{Evaluation metrics} \label{subsec:metric}
As briefly reviewed in \cref{subsec:metricreview}, proper evaluation criteria for 3D generative models are still in active development. The shortcomings of commonly used COV and MMD metrics \cite{achlioptas2018learning} were verified by \cite{yang2019pointflow,ibing20213d}, and replaced by 1-NNA~\cite{yang2019pointflow} or ECD~\cite{ibing20213d}. We follow their suggestions and use 1-NNA and ECD as two of our evaluation metrics and also report COV and MMD as a reference. The definitions of these metrics are at below.

Let $S_g\sim \mathbb{P}_g$ be the set of generated samples, $S_r \sim \mathbb{P}_r$ be the set of reference data, and $D(\cdot,\cdot): \mathcal{X}\times\mathcal{X}\mapsto \mathbb{R}_{\geq 0}$ be the distance function.

\myparagraph{COV~\cite{achlioptas2018learning}}
For any $X\in S_g$, its nearest neighbor $Y\in S_r$ is marked as a \emph{match}, and COV measures the fraction of $Y\in S_r$ matched to any element in $S_g$:
\begin{equation}
    \operatorname{COV}(S_g, S_r)=\frac{|\{\operatorname{argmin}_{Y \in S_{r}} D(X, Y) \mid \forall X \in S_{g}\}|}{|S_{r}|}.
    \label{eq:cov_def}
\end{equation}

\myparagraph{MMD~\cite{achlioptas2018learning}}
MMD measures the average distance from any $Y\in S_r$ to its nearest neighbor $X\in S_g$:
\begin{equation}
    \operatorname{MMD}(S_g, S_r)=\frac{1}{|S_{r}|} \sum_{Y \in S_{r}} \min _{X \in S_{g}} D(X, Y).
    \label{eq:mmd_def}
\end{equation}

\myparagraph{1-NNA~\cite{yang2019pointflow}}
Let $S_{-X} = S_r \cup S_g - \{X\}$ and $N_X$ be the nearest neighbor to $X$ in $S_{-X}$. 1-NNA is the leave-one-out accuracy of the 1-NN classifier:
\begin{equation}
    \text{1-NNA} (S_{g}, S_{r})=\frac{\sum_{X \in S_{g}} \mathbb{I}[N_{X} \in S_{g}]+\sum_{Y \in S_{r}} \mathbb{I}[N_{Y} \in S_{r}]}{|S_{g}|+|S_{r}|},
\end{equation}
where $\mathbb{I}[\cdot]$ is the indicator function.

\myparagraph{ECD~\cite{ibing20213d}} A $k$-minimum spanning tree of the neighborhood graph of $S_r \cup S_g$ is built. Three types of tree edges are defined: connected within $S_r$,  connected within $S_g$, connected between $S_r$ and $S_g$. ECD is the weighted difference between the number of these edges and the edge count if $S_r$ and $S_g$ are from the same distribution. The exact formula of ECD can be found in the Appendix of \cite{ibing20213d}.

For COV, MMD, LFD and ECD computations, we follow \cite{chen2019learning,ibing20213d} to use mesh-based light-field-distance (LFD) as the distance function $D$, as our method and all the compared methods can extract mesh surfaces for evaluation.

\begin{figure}[t]
    \centering
    \includegraphics[width=1\linewidth]{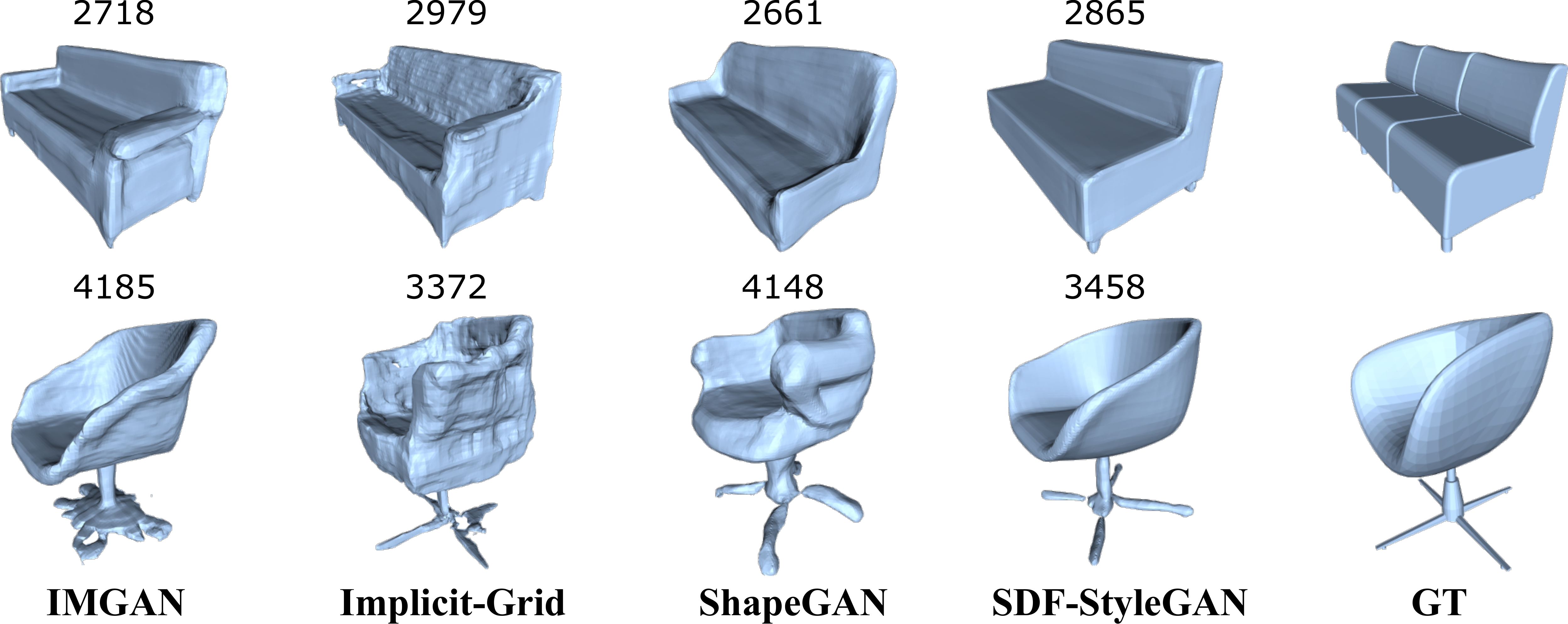}
    \caption{Illustration of the drawback of LFD.  The number above the shape is the LFD between the shape and its GT counterpart.}
    \label{fig:lfdissue} \vspace{-8mm}
\end{figure}

\myparagraph{Drawback of LFD} Although the above metrics based on LFD is recommended by previous 3D generation works~\cite{chen2019learning,ibing20213d},  a smaller LFD between shape $A$ and shape $B$ does not mean that their visual similarity is better than that of shape $A$ and another shape $C$ with a larger LFD, as LFD is based on silhouette images without considering the fidelity of the local shape geometry.
We illustrate this drawback in \cref{fig:lfdissue} as follows. The shape in the rightmost column is sampled from ShapeNet, denoted by GT. We select the shape with the smallest LFD value to the GT shape for each method. The numbers above these shapes are the LFD values. We can see that some shapes with bumpy geometry have smaller LFDs although they have a large visual difference from their GT counterparts.  This drawback indicates that the LFD-based evaluation metrics for 3D shape generative networks are not sufficient to measure the visual and geometry quality of generated shapes.

\begin{figure}[t]
    \centering
    \includegraphics[width=\linewidth]{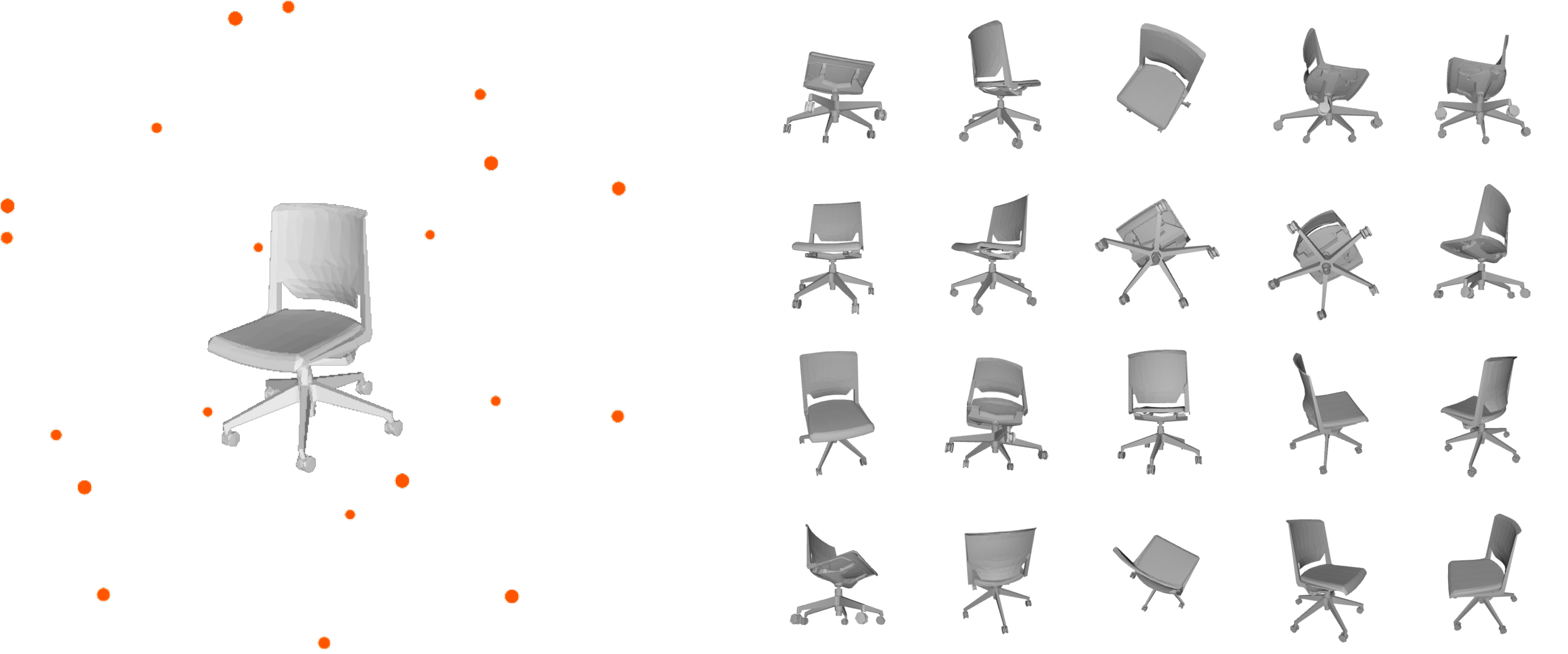}
    \caption{Left: configuration of rendering views. 20 camera positions are illustrated as red points. Right: 20 rendered images for computing FID, the image resolution is $299\times299$.}
    \label{fig:camera} \vspace{-7mm}
\end{figure}

\myparagraph{Shading-image-based FID} To resolve the drawback of using LFD and take human perception into consideration, we propose to adapt Fr\'echet inception distance (FID)~\cite{heusel2017gans} on the shading images of 3D shapes, as the visual quality of 3D shapes for humans is mostly perceived from rendered view images. We name this metric by shading-image-based FID. We normalize the surface mesh into a unit sphere and render shading images from 20 uniformly distributed views, here the view selection is the same as the LFD algorithm~\cite{chen2003visual} (see the illustration in \cref{fig:camera}). For the generated data set and the training set, we compute the Fr\'echet inception distance (FID) score based on their $i$-th view images, and average 20 FID scores to define the shading-image-based FID:
\begin{equation}
    FID = \frac{1}{20}\left[\sum_{i=1}^{20} \|\mu_{g}^i - \mu_{r}^i\|^2 + \operatorname{Tr}\left(\Sigma_{g}^i + \Sigma_{r}^i - 2 \left(\Sigma_{r}^i\Sigma_{g}^i\right)^{1/2}\right)\right],
\end{equation}
where $g$ and $r$ denote the features of the generated data set and the training set, $\mu_{\cdot}^i$, $\Sigma_{\cdot}^i$ denote the mean and the covariance matrix of the corresponding shading images rendered from the $i$-th view.

Here, although the rendered images of 3D shapes are different from natural images in ImageNet~\cite{deng2009imagenet} trained for Inception-V3~\cite{szegedy2016rethinking}, we found that the shading-image-based FID is meaningful, and the generated data set with a smaller FID has more plausible and visually similar shapes to the training set.

\myparagraph{FPD~\cite{shu20193d}} We also adapt Fr\'echet point cloud distance (FPD) proposed by \cite{shu20193d} for evaluation. For any shape in $S_g$ or $S_r$, we sample 2048 points on the mesh surface and pass them to a DGCNN backbone network~\cite{wang2019dynamic,li2021sp} pre-trained with the shape classification task. The output feature vectors of $S_g$ and $S_r$ are used to compute FPD. Small FPD values are better.

\myparagraph{Evaluation setup} For computing COV and MMD, we follow the setting of \cite{chen2019learning}: $S_r$ is formed by the original ShapeNet meshes in the test dataset, $5|S_r|$ shapes are generated as $S_g$. For computing 1-NNA and ECD, $S_r$ is not changed and $|S_r|$ shapes are sub-sampled from $S_g$ as $S_g^\prime$. 1-NNA and ECD are evaluated 10 times between $S_g^\prime$ and $S_r$, and their average numbers are reported.  We note that taking the training set as $S_r$ is more reasonable to evaluate the capability of GANs since the distribution of the test set could be very different with the training set and the number of shapes in the test set is usually small. However, here we still follow the setup of the original papers for consistency.

For computing shading-image-based FID and FPD, $S_r$ is formed by the meshes in the training set and $|S_r|$ shapes are generated as $S_g$. We employ the clean-fid algorithm~\cite{parmar2021cleanfid} to calculate FID. The default iso-values of the competitive methods are employed to extract surface meshes by the Marching Cube algorithm~\cite{lorensen1987marching} on the $128^3$ grids.

\begin{table}[t]
    \caption{Quantitative evaluation of different methods. COV, MMD, 1-NNA and ECD use Light-field-distance and their reference set $S_r$ is the test set of shape categories. FID is the abbreviation of shading-image-based FID.  }
    \label{tab:comp-tab}
    \centering
    \resizebox{\columnwidth}{!}
    {
        \begin{tabular}{c|c|*{6}{r}}
            \toprule
            \thead{Data}                                        & \thead{Method} & \thead{COV(\%) $\uparrow$} & \thead{MMD $\downarrow$} & \thead{1-NNA$\downarrow$} & \thead{ECD$\downarrow$} & \thead{FID$\downarrow$} & \thead{FPD$\downarrow$} \\
            \midrule
            \multirow{4}{*}{\rotatebox[origin=c]{90}{Chair}}    & IMGAN          & 72.57                      & \textbf{3326}            & 0.7042                    & 1998                    & 63.42                   & 1.093                   \\

                                                                & Implicit-Grid  & \textbf{82.23}             & 3447                     & \textbf{0.6655}           & \textbf{1231}           & 119.5                   & 1.456                   \\
                                                                & ShapeGAN       & 65.19                      & 3726                     & 0.7896                    & 4171                    & 126.7                   & 1.177                   \\
                                                                & SDF-StyleGAN   & 75.07                      & 3465                     & 0.6690                    & 1394                    & \textbf{36.48}          & \textbf{1.040}          \\
            \midrule
            \multirow{4}{*}{\rotatebox[origin=c]{90}{Airplane}} & IMGAN          & 76.89                      & \textbf{4557}            & \textbf{0.7932}           & \textbf{2222}           & 74.57                   & 1.207                   \\

                                                                & Implicit-Grid  & \textbf{81.71}             & 5504                     & 0.8509                    & 4254                    & 145.4                   & 2.341                   \\
                                                                & ShapeGAN       & 60.94                      & 5306                     & 0.8807                    & 6769                    & 162.4                   & 2.235                   \\
                                                                & SDF-StyleGAN   & 74.17                      & 4989                     & 0.8430                    & 3438                    & \textbf{65.77}          & \textbf{0.942}          \\

            \midrule
            \multirow{4}{*}{\rotatebox[origin=c]{90}{Car}}      & IMGAN          & 54.13                      & 2543                     & 0.8970                    & 12675                   & 141.2                   & 1.391                   \\
                                                                & Implicit-Grid  & \textbf{75.13}             & 2549                     & 0.8637                    & 8670                    & 209.3                   & 1.416                   \\
                                                                & ShapeGAN       & 57.40                      & 2625                     & 0.9168                    & 14400                   & 225.2                   & 0.787                   \\
                                                                & SDF-StyleGAN   & 73.60                      & \textbf{2517}            & \textbf{0.8438}           & \textbf{6653}           & \textbf{97.99}          & \textbf{0.767}          \\
            \midrule
            \multirow{4}{*}{\rotatebox[origin=c]{90}{Table}}    & IMGAN          & 83.43                      & \textbf{3012}            & \textbf{0.6236}           & \textbf{907}            & 51.70                   & 1.022                   \\

                                                                & Implicit-Grid  & \textbf{85.66}             & 3082                     & 0.6318                    & 1089                    & 87.69                   & 1.516                   \\
                                                                & ShapeGAN       & 76.26                      & 3236                     & 0.7069                    & 1913                    & 103.1                   & \textbf{0.934}          \\
                                                                & SDF-StyleGAN   & 69.80                      & 3119                     & 0.6692                    & 1729                    & \textbf{39.03}          & 1.061                   \\
            \midrule
            \multirow{4}{*}{\rotatebox[origin=c]{90}{Rifle}}    & IMGAN          & 71.16                      & \textbf{5834}            & 0.6911                    & 701                     & 103.3                   & 2.102                   \\
                                                                & Implicit-Grid  & 77.89                      & 5921                     & \textbf{0.6648}           & \textbf{357}            & 125.4                   & 1.904                   \\
                                                                & ShapeGAN       & 46.74                      & 6450                     & 0.8446                    & 3115                    & 182.3                   & 1.249                   \\
                                                                & SDF-StyleGAN   & \textbf{80.63}             & 6091                     & 0.7180                    & 510                     & \textbf{64.86}          & \textbf{0.978}          \\
            \bottomrule
        \end{tabular}
    }
    \vspace{-4mm}
\end{table}

\subsection{Quantitative and qualitative evaluation}\label{subsec:comp}

\myparagraph{Quantitative evaluation} We evaluate SDF-StyleGAN, IMGAN, ShapeGAN and Implicit-Grid by using the metrics listed in \cref{subsec:metric}. Quantitative results are provided in \cref{tab:comp-tab}. LFD-based COV and MMD are listed for reference only as mentioned above.  In the categories of airplane and table, IMGAN has better 1-NNA and ECD values. In the categories of chair and rifle, Implicit-Grid has better performance in 1-NNA and ECD. Our SDF-StyleGAN achieves the smallest 1-NNA and ECD in car shapes. However, the 1-NNA and ECD metrics do not faithfully reflect how the distribution of the generated set is similar to that of the training set, as the reference set is the test set. In terms of FID and FPD that use the training set as the reference set, our SDF-StyleGAN is significantly better than other methods, while IMGAN is the second best.

\begin{figure*}[t]
    \centering
    \includegraphics[width=\linewidth]{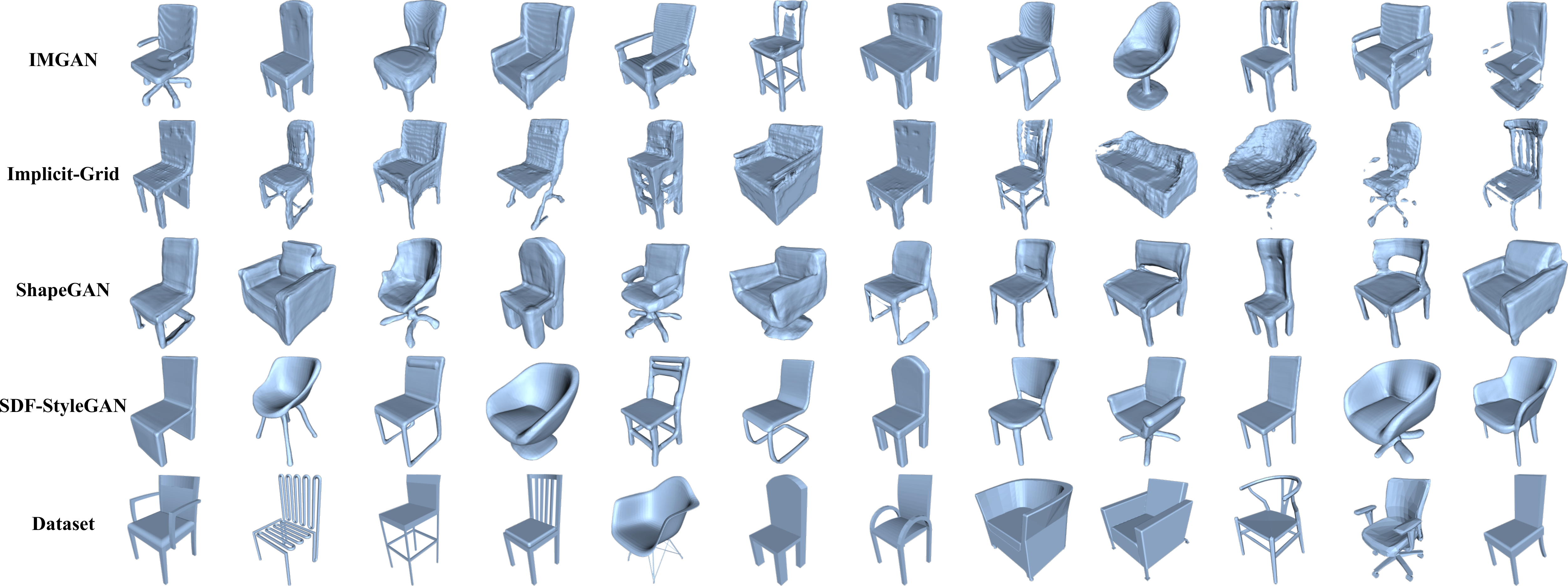}
    \caption{Visual comparison of randomly generated chairs by different methods. The shapes in the last row are randomly sampled from the training dataset.}
    \label{fig:chaircomp} \vspace{-2mm}
\end{figure*}

\begin{figure*}[t]
    \centering
    \includegraphics[width=\linewidth]{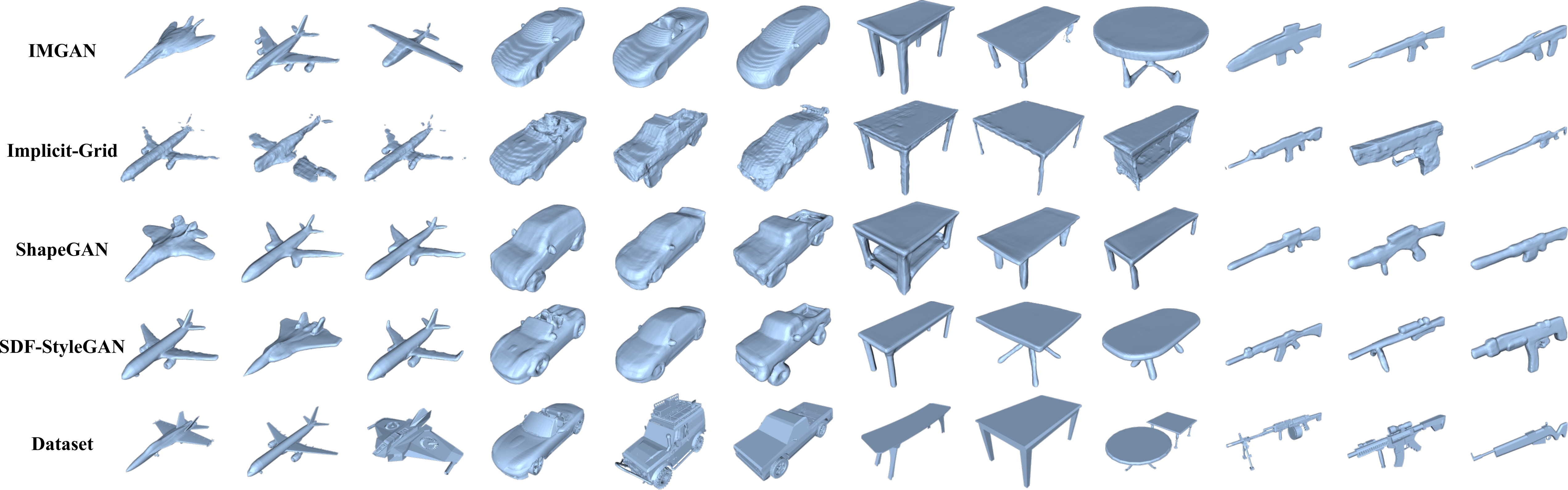}
    \caption{Visual comparison of randomly generated shapes by different methods on airplane, car, table and rifle. The shapes in the last row are randomly sampled from the dataset.}
    \label{fig:otherclass} \vspace{-5mm}
\end{figure*}

\myparagraph{Qualitative evaluation} The superiority of SDF-StyleGAN in terms of FID and FPD metrics can be perceived directly through visual comparison. In \cref{fig:chaircomp}, a set of chairs generated by each method is illustrated for comparison. For IMGAN, their chairs have plausible structures, but with bumpy geometry, probably because IMGAN learned the voxelization artifacts in constructing the occupancy field for the training data. For Implicit-Grid, its results contain many missing regions and are also bumpy. The visual quality of its results is the worst, although it has the best 1-NNA and ECD values among all methods. The ShapeGAN results do not have voxelization-like bump geometry, but the shapes are distorted compared to the training data.  Our SDF-StyleGAN generates more geometry plausible chair: the seats are flat, the local geometry is smooth, and the overall structure is more complete. This significant improvement in geometry and visual quality by our method is correctly reflected by the FID values in \cref{tab:comp-tab}.  In \cref{fig:otherclass}, we also show the generated shapes in other shape categories, and a similar conclusion remains. In our supplemental material, we provide more randomly generated results (240 shapes for each category) without any cherry picking. These results further validate the capability and superiority of our method.


\subsection{Ablation studies} \label{subsec:ablation}
We conducted a series of ablation studies to validate our framework design, using the chair category from the ShapeNet dataset.

\myparagraph{Efficacy of local discriminator and SDF gradients} Five alternative configurations on using the global discriminator $\mathcal{D}_G$ and the local discriminator $\mathcal{D}_L$ were tested:
\begin{enumerate}[label=(\arabic*)]
    \item\label{item:ablation_1} use $\mathcal{D}_G$ only, and use the SDF values as input;
    \item \label{item:ablation_2} use $\mathcal{D}_G$ only, and use both SDF values and gradients as input;
          \item\label{item:ablation_3} add $\mathcal{D}_L$ to \ref{item:ablation_2}, and use SDF values as input to $\mathcal{D}_L$;
          \item\label{item:ablation_4} add $\mathcal{D}_L$ to \ref{item:ablation_2}, and use SDF gradients as input to $\mathcal{D}_L$;
          \item\label{item:ablation_5} our default configuration that uses both $\mathcal{D}_G$ and $\mathcal{D}_L$ and feeds both SDF and SDF gradients to $\mathcal{D}_G$ and $\mathcal{D}_L$.
\end{enumerate}

\begin{figure}[t]
    \centering
    \begin{overpic}[width=0.95\linewidth]{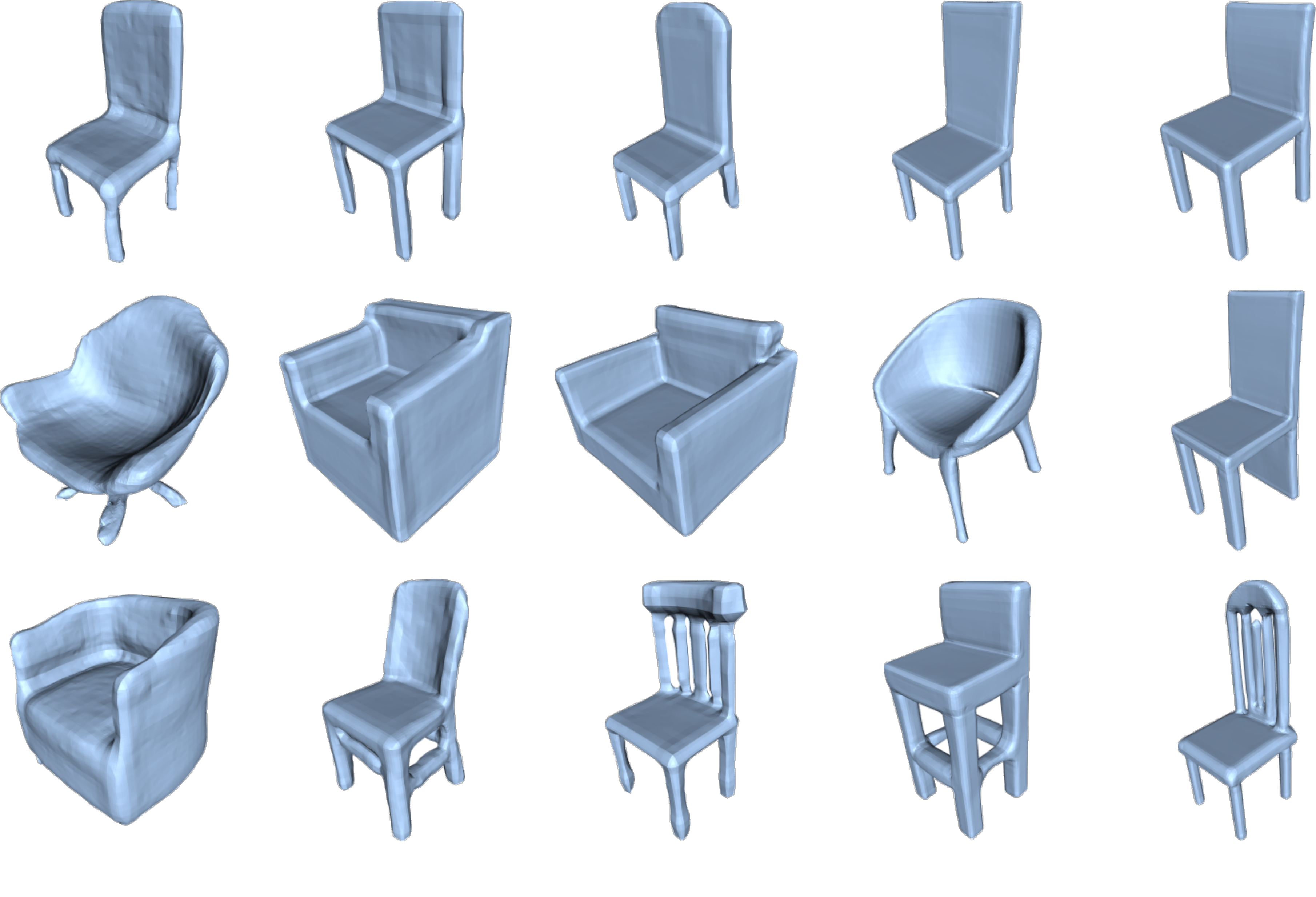}
        \put(7,0){\ref{item:ablation_1}}
        \put(28,0){\ref{item:ablation_2}}
        \put(49.5,0){\ref{item:ablation_3}}
        \put(71,0){\ref{item:ablation_4}}
        \put(92.5,0){\ref{item:ablation_5}}
    \end{overpic}
    \caption{Visual illustration of the generated shapes by SDF-StyleGAN trained with different configurations listed in \cref{tab:abalation-tab1}. In the first row we intentionally pick a four-leg chair from each configuration for visual comparison. The shapes in the other rows are randomly selected.}
    \label{fig:ablation} \vspace{-4mm}
\end{figure}

\cref{tab:abalation-tab1} shows the evaluation metrics of SDF-StyleGAN trained with these configurations. By comparing \ref{item:ablation_3}, \ref{item:ablation_4} and \ref{item:ablation_5} with \ref{item:ablation_1} and \ref{item:ablation_2}, we can see that the addition of $\mathcal{D}_L$ increases the performance significantly; by comparing \ref{item:ablation_2} with \ref{item:ablation_1}, and \ref{item:ablation_4} with \ref{item:ablation_3}, we can find that the use of SDF gradients increases performance by a large margin. In \cref{fig:ablation}, we visualize some shapes generated by the networks corresponding to these five configurations. We can clearly see that networks with the local discriminator produce more smooth and regular shapes, and the use of SDF gradients effectively reduces geometry distortion.

\begin{table}[t]
    \caption{Ablation study on the use of local discriminator and SDF gradients.  $G$ and $L$ denotes the global discriminator and the local discriminator, respectively. $0$ and $1$ denotes the use of SDF values and the use of SDF gradients, respectively. The combinations of $G,L$ and $0,1$ form the five configurations.
    }
    \label{tab:abalation-tab1}
    \centering
    \resizebox{0.95\columnwidth}{!}{
        \begin{tabular}{c|cccc|cccccc}
            \toprule
            \thead{Config.}       & \thead{G0} & \thead{G1} & \thead{L0} & \thead{L1} & \thead{1-NNA$\downarrow$} & \thead{ECD$\downarrow$} & \thead{FID$\downarrow$} & \thead{FPD$\downarrow$} \\
            \midrule
            \ref{item:ablation_1} & \ding{51}  &            &            &            & 0.8339                    & 6280                    & 173.3                   & 1.335                   \\
            \ref{item:ablation_2} & \ding{51}  & \ding{51}  &            &            & 0.7642                    & 3387                    & 85.46                   & 1.198                   \\
            \ref{item:ablation_3} & \ding{51}  & \ding{51}  & \ding{51}  &            & 0.7124                    & 2436                    & 92.35                   & 1.111                   \\
            \ref{item:ablation_4} & \ding{51}  & \ding{51}  &            & \ding{51}  & 0.6708                    & 1541                    & 39.17                   & 1.123                   \\
            \ref{item:ablation_5} & \ding{51}  & \ding{51}  & \ding{51}  & \ding{51}  & \textbf{0.6690}           & \textbf{1394}           & \textbf{36.48}          & \textbf{1.040}          \\
            \bottomrule
        \end{tabular}
    }
    \vspace{-4mm}
\end{table}

\myparagraph{Candidate local region number $N_0$} Local region selection is important to our training. A smaller $N_0$ does not help to distribute the selected local regions more evenly, while a much larger $N_0$ cannot ensure that the selected local regions are around the zero isosurface. We tested three choices of $N_0$: 512, 2048, 8912, and found that the default value $2048$ can lead to better performance, as shown in \cref{tab:abalation-tab2}.

\begin{table}[t]
    \caption{Ablation study on the number of candidate local regions.}
    \label{tab:abalation-tab2}
    \centering
    \resizebox{0.7\columnwidth}{!}
    {
        \begin{tabular}{c|cccc}
            \toprule
            \thead{$N_0$} & \thead{1-NNA$\downarrow$} & \thead{ECD$\downarrow$} & \thead{FID$\downarrow$} & \thead{FPD$\downarrow$} \\
            \midrule
            512           & 0.8875                    & 4249                    & 85.90                   & 2.597                   \\
            2048          & \textbf{0.6690}           & \textbf{1394}           & \textbf{36.48}          & \textbf{1.040}          \\
            8192          & 0.6881                    & 1870                    & 45.37                   & 1.108                   \\
            \bottomrule
        \end{tabular}
    }
\end{table}

\begin{figure}[t]
    \centering
    \begin{overpic}[width=0.8\linewidth]{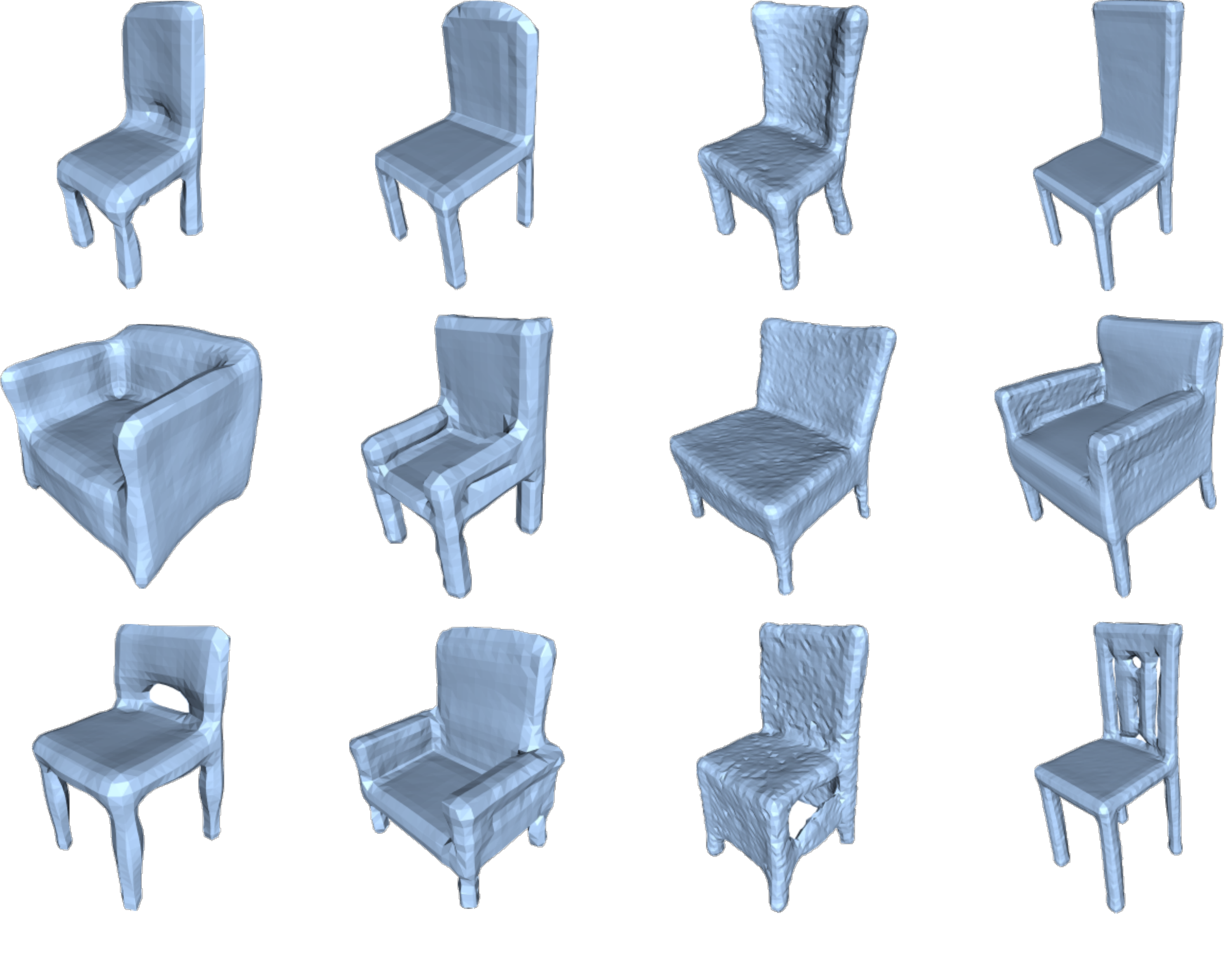}
        \put(7,0){\tiny $N=32$}
        \put(27,0){\tiny $N=32$ / with grad.}
        \put(59,0){\tiny $N=64$}
        \put(81,0){\tiny $N=64$ / with grad.}
    \end{overpic}
    \caption{Illustration of the randomly generated shapes by the alternative design of SDF-StyleGAN that generates discrete SDF grids directly. "with grad." means the SDF gradients also be fed into the discriminator.}
    \label{fig:nomlp} \vspace{-5mm}
\end{figure}

\myparagraph{Feature-volume-based implicit SDF} Our method benefits from feature-volume-based implicit SDF. We did an ablation study by replacing it with discrete SDFs, \ie letting the generator directly output an $N\times N \times N$ SDF grid. We use the global discriminator only for this test. We tested two kinds of resolutions: $N=32$ and $N=64$, and we also tested whether the additional SDF gradient input can help improve these networks. \cref{tab:abalation-tab4} shows the performance of these alternative networks, where the mesh extraction by the Marching Cube algorithm uses the same resolution of the SDF grid. We found that the use of SDF gradients can significantly improve the performance of these alternate networks, but there is still a large performance gap between them and our default design. \cref{fig:nomlp} illustrates the shapes generated by these alternative networks.

\begin{table}[t]
    \caption{Performance of an alternative network design that generates discrete SDF grids directly.}
    \label{tab:abalation-tab4}
    \centering
    \resizebox{0.7\columnwidth}{!}
    {
        \begin{tabular}{cc|cc}
            \toprule
            \thead{resolution} & \thead{SDF gradient} & \thead{FID$\downarrow$} & \thead{FPD$\downarrow$} \\
            \midrule
            $32^3$             & \ding{55}            & 156.0                   & 2.097                   \\
            $32^3$             & \ding{51}            & 104.6                   & 1.532                   \\
            $64^3$             & \ding{55}            & 154.2                   & 1.323                   \\
            $64^3$             & \ding{51}            & 80.26                   & 1.792                   \\

            \bottomrule
        \end{tabular}
    }
\end{table}

\section{Applications} \label{sec:app}

Based on the GAN inversion technique, we employed the trained SDF-StyleGAN generator for a series of applications.

\myparagraph{3D GAN inversion}
The goal of 3D GAN inversion is to embed a shape into the latent space of GAN. To this end, we first use an encoder network $\mathcal{E}_\varphi$ to encode the input sample $\mat{x}$ to a latent code $\mat{z} \in \mathcal{Z}$ or $\mat{w} \in \mathcal{W}$, then feed the resulting code to our trained SDF-StyleGAN model $\mathcal{F}_\vartheta$, and optimize the parameters $\varphi$ with the following loss function:
\begin{equation}
    \mathcal{L_{\varphi}} = \frac{1}{|\mathcal{V}|} \sum_{\mp \in \mathcal{V} }
    \left| \mathcal{F}_\vartheta \left(\mathcal{E}_\varphi \left(\mat{x}\right); \mp \right) - \texttt{SDF}_{gt}\left(\mp\right) \right|,
    \label{eq:inversion}
\end{equation}
where $\mathcal{V}$ is a set of points sampled inside a volumetric space that contains $\mat{x}$, $\texttt{SDF}_{gt}$ queries the ground-truth SDF values of any point with respect to $\mat{x}$, and $\mathcal{F}_\vartheta\left(\mat{y}; \mp\right)$ returns the predicted SDF value at point $\mp$ for a given latent code $\mat{y}$.
After training, we can directly map $\mat{x}$ to a latent code $\mat{y}$ via network forwarding.
The choice of encoders is flexible and depends on the input type of $\mat{x}$.

\myparagraph{Shape reconstruction from point clouds}
We use SDF-StyleGAN to reconstruct category-wise shapes from point clouds. We first train the encoder of 3D GAN inversion on a shape category. We choose a light version of octree-based CNN~\cite{Wang2017}, which is composed of 3 sparse convolution layers and 2 fully connected layers. We optimize $\varphi$ with an AdamW optimizer~\cite{loshchilov2018decoupled} for 200 epochs on a shape collection while keeping $\vartheta$ fixed.
We use the 3D GAN inversion to map an input point cloud to an SDF-StyleGAN latent code via the trained encoder.
The latent code can be further optimized by minimizing the following loss function which forces the SDF values at $\mp$ to be zero:
\begin{equation}
    \mathcal{L_{\mat{y}}} = \frac{1}{|\mathcal{P}|} \sum_{\mp \in \mathcal{P}}
    \left|\mathcal{F}_\vartheta \left( \mat{y}; \mp \right)) \right|,
    \label{eq:finetune}
\end{equation}
where $\mathcal{P}$ is a set of points sampled on the surface of $\mat{x}$.
In our experiment, we optimize the code obtained from the encoder with 1000 iterations for each input point cloud $\mat{x}$.

\begin{figure}[t]
    \centering
    \begin{overpic}[width=\linewidth]{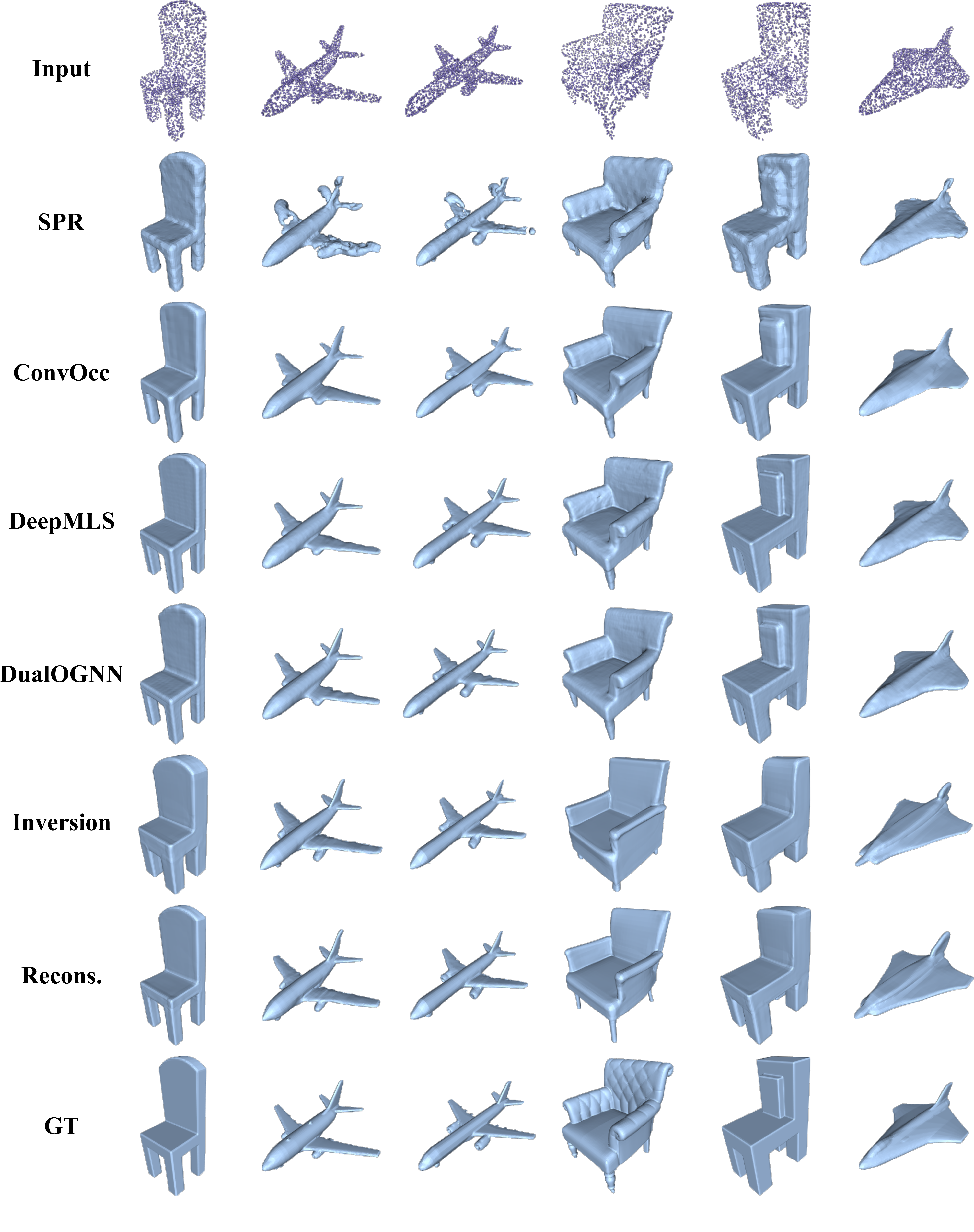}
        \put(12.5,0){(a)}
        \put(23.5,0){(b)}
        \put(36.5,0){(c)}
        \put(48.5,0){(d)}
        \put(61,0){(e)}
        \put(73,0){(f)}
    \end{overpic}
    \caption{Shape reconstruction. Inversion and Recons. denote the reconstructed shapes via our GAN inversion and our reconstruction, respectively. The shapes in GT row are the ground-truth shapes. Here the point normals requested by SPR~\cite{kazhdan2013screened} are estimated from 10 nearest points.}
    \label{fig:recon} \vspace{-6mm}
\end{figure}

\cref{fig:recon} shows some reconstructed shapes using our method, as well as other state-of-the-art methods designed for surface reconstruction, including Screen Poisson reconstruction (SPR)~\cite{kazhdan2013screened}, and three learning-based methods: ConvOcc~\cite{peng2020convolutional}, DeepMLS~\cite{liu2021deep}, and DualOGNN~\cite{Wang2022dualocnn}). Each input noisy point cloud contains 3000 points, where the noisy level is the same as the noisy data used in ConvOcc, DeepMLS and DualOGNN. As our latent code is constrained by the SDF-StyleGAN space, the geometry of the reconstructed shapes respects the data distribution of the training set of SDF-StyleGAN. We can find that our approach is robust to noise, but has limitations in fitting unusual shape details, such as the bumpy surface of the backrest (\cref{fig:recon}-(d)), the additional backrest (\cref{fig:recon}-(e)), and the cabin of the fighter plane  (\cref{fig:recon}-(f)).

\begin{figure}[t]
    \centering
    \begin{overpic}[width=\linewidth]{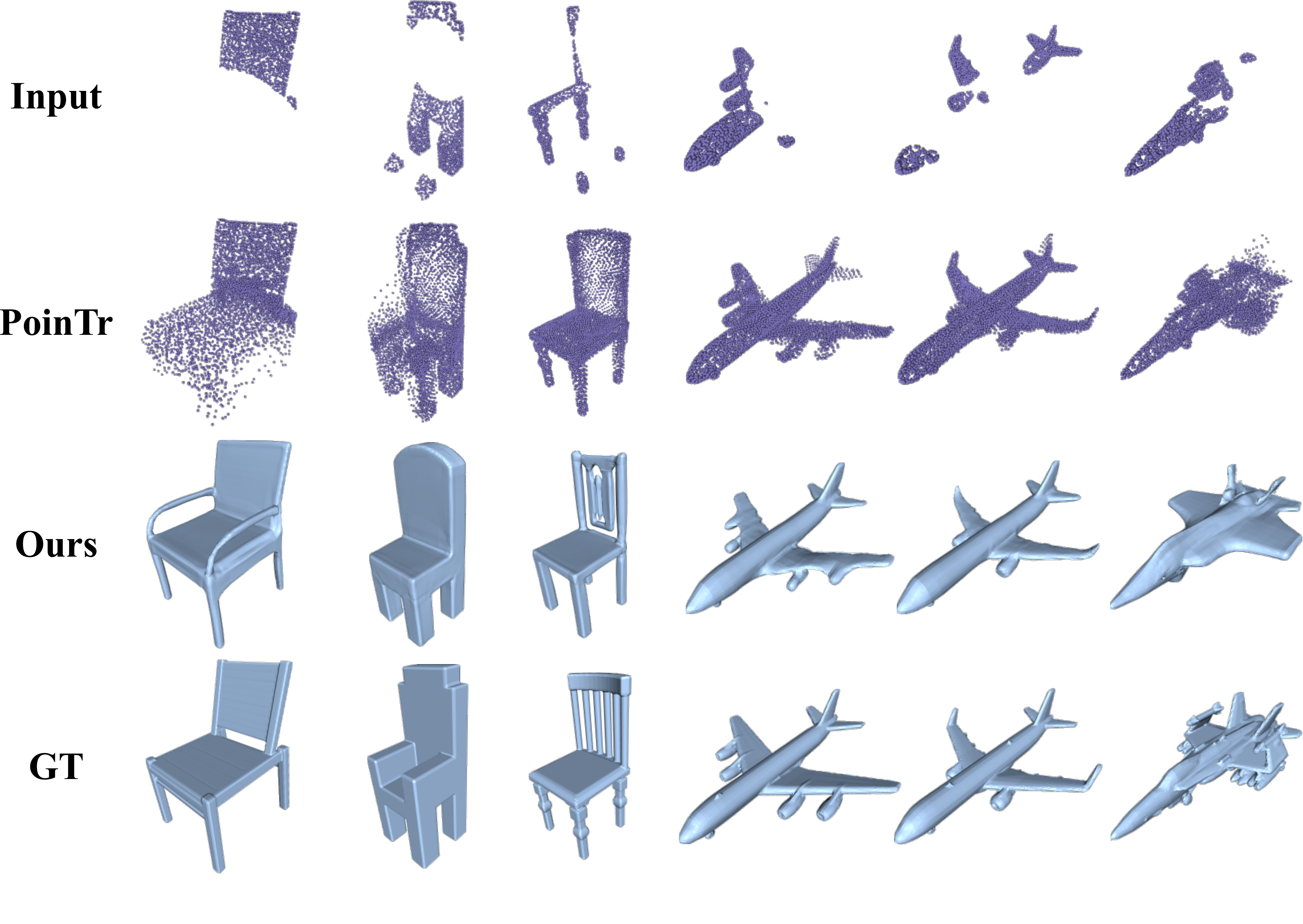}
        \put(15,0){(a)\label{item:fig:completion:col1}}
        \put(29,0){(b)\label{item:fig:completion:col2}}
        \put(43,0){(c)\label{item:fig:completion:col3}}
        \put(58,0){(d)\label{item:fig:completion:col4}}
        \put(74.5,0){(e)\label{item:fig:completion:col5}}
        \put(91,0){(f)\label{item:fig:completion:col6}}
    \end{overpic}
    \caption{Shape completion. The shapes in GT row are the ground-truth shapes of the input point clouds. Our method can recover plausible shapes from incomplete inputs.}
    \label{fig:completion} \vspace{-2mm}
\end{figure}

\myparagraph{Shape completion}
It is easy to adapt the 3D GAN inversion for the shape completion task.
In this experiment, we randomly drop 75\% of the input point cloud $\mat{x}$ to train the encoder while keeping the loss functions unchanged. The trained encoder maps the incomplete input to a latent code that always corresponds to a plausible shape due to the GAN training. We can further optimize the latent code via minimizing \cref{eq:finetune}. In \cref{fig:completion}, we test our method on the categories of chairs and airplanes and compare it with a state-of-the-art point completion cloud approach PoinTr~\cite{yu2021pointr}, which is based on a transformer architecture. We can see that within the shape space constrained by SDF-StyleGAN, our method has a better ability to reconstruct plausible shapes, even from a partial chair back (see \cref{fig:completion}-(a)). However, similar to the behavior on shape reconstruction, our method has worse capability in fitting fine details, such as the round joint of chair legs (see \cref{fig:completion}-(c)) and the wingtip of the airplane (see \cref{fig:completion}-(f)); while PoinTr keeps the original points to maintain the input geometry.

\begin{figure}[t]
    \centering
    \includegraphics[width=\linewidth]{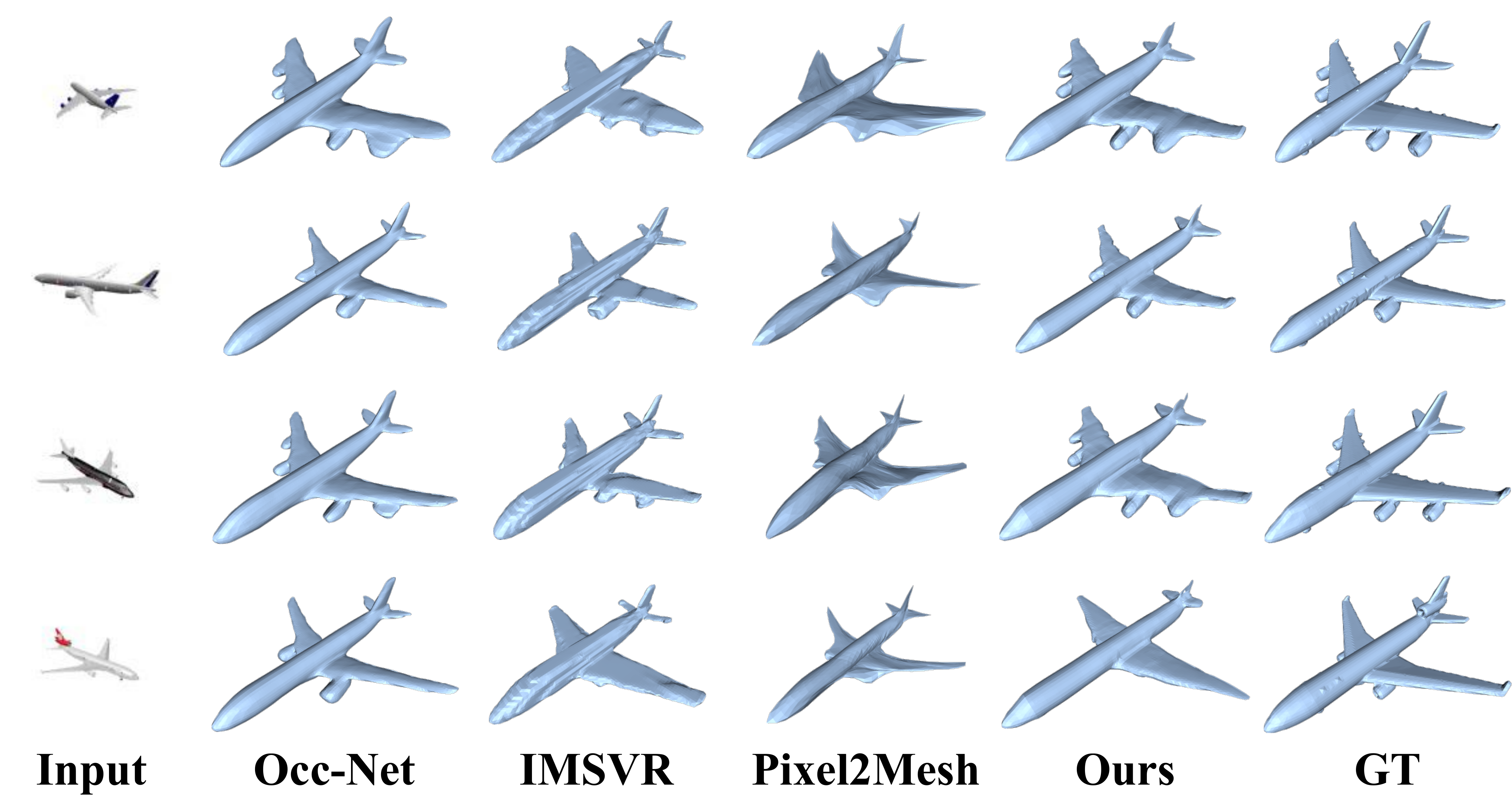}
    \caption{Shape generation from single images. The shapes in GT column are the ground-truth shapes.} 
    \label{fig:svr} \vspace{-4mm}
\end{figure}

\myparagraph{Shape generation from single images} It is also easy to adapt the GAN inversion to predict a 3D shape from a single image input. In this experiment, the input sample $\mat{x}$ is an image, and we use a ResNet-18 network~\cite{he2016deep} to encode the input. We trained the encoder network on the airplane category, where the low-resolution training images are from
\cite{choy20163d}. \cref{fig:svr} shows some results generated by our method, as well as some state-of-the-art methods including IMSVR~\cite{chen2019learning}, Pixel2Mesh~\cite{wang2018pixel2mesh}, Occ-Net~\cite{mescheder2019occupancy}. Our method is able to predict plausible shapes from low-resolution images. The result in the last row shows a failure case that our method does not generate the airplane engine, as the encoder does not map the input to a good latent code.

\begin{figure}[t]
    \centering
    \includegraphics[width=\linewidth]{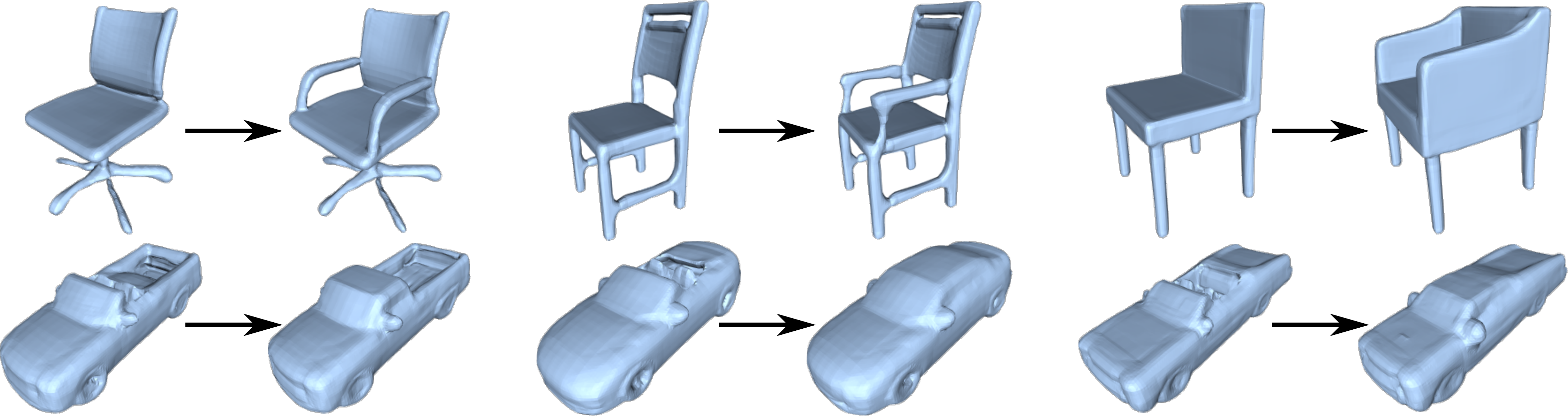}
    \caption{Shape style editing based on the algorithm of \cite{shen2020interfacegan}. The style editing on chairs is to add arms, and the style editing on cars is to add car roofs. }
    \label{fig:interfacegan} \vspace{-2mm}
\end{figure}

\myparagraph{Shape style editing}
To edit the attributes of facial images via GAN model, Shen \etal~\cite{shen2020interfacegan} assumed that for any binary semantics, there may exist a hyperplane in the latent space serving as the separation boundary. This hyperplane in the latent space can be used for editing the image style. We adapted their approach to our SDF-StyleGAN space for 3D shape style editing.  We use the shape labels provided in ~\cite{muralikrishnan2018tags2parts} to divide the chair category into two groups: chairs with and without arms. A DGCNN network \cite{wang2019dynamic} was trained to infer whether an input chair contains arms or not. We then randomly generated 10000 chairs using SDF-StyleGAN and divided them into two groups using the trained classification network. As the latent codes of these generate shapes are known, the hyperplane $\mat{n} \cdot \mat{x} + d = 0$ that separates these two groups can easily be computed. For any generated shape or a shape obtained by GAN inversion, we can change its latent code $\mat{y}$ to $\mat{y} - \eta (\mat{n} \cdot \mat{y} + d) \mat{n}$, $\eta \geq 0$, to modify the shape style from ``without arms'' to ``with arms'', or vice versa. Similarly on the car category, we use the label ``with roof'' and ``without roof'' to find the separation plane and edit the shape style. In \cref{fig:interfacegan}, we demonstrate this kind of shape style editing, and we find that the original geometry of the input shapes is also well-preserved.

\begin{figure}[t]
    \centering
    \includegraphics[width=\linewidth]{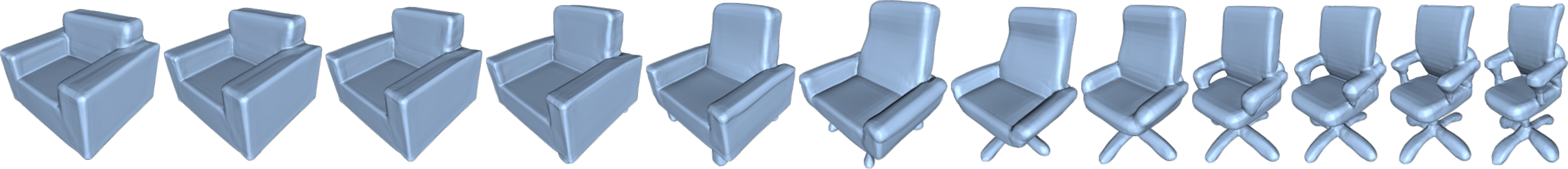}
    \caption{An example of shape interpolation in the $\mathcal{Z}$ space.}
    \label{fig:interpolation} \vspace{-4mm}
\end{figure}

\myparagraph{Shape interpolation} A straightforward application is to perform shape interpolation in the $\mathcal{Z}$ space, or the $\mathcal{W}$ space. \cref{fig:interpolation} shows the intermediate results of the interpolation between two chairs, in the $\mathcal{Z}$ space. Due to the use of GAN model, each intermediate result is plausible, and the transition of shape geometry between adjacent frames is also smooth.

\section{Conclusion} \label{sec:conclusion}

In the presented work, we offer a high-quality 3D generative model --- SDF-StyleGAN for shape generation, which is capable of producing diverse and visually plausible 3D models superior to the state-of-the-art. These significant improvements are ascribed to the design of our global and local SDF discriminators, the choice of implicit SDF representation, the use of SDF gradients, and the StyleGAN2 network structure.  We evaluated the learned generative model using suitable 3D GAN metrics, including our proposed FID on rendered images, and demonstrated the capability of SDF-StyleGAN on a series of applications.

\begin{figure}[t]
    \centering
    \begin{overpic}[width=\linewidth]{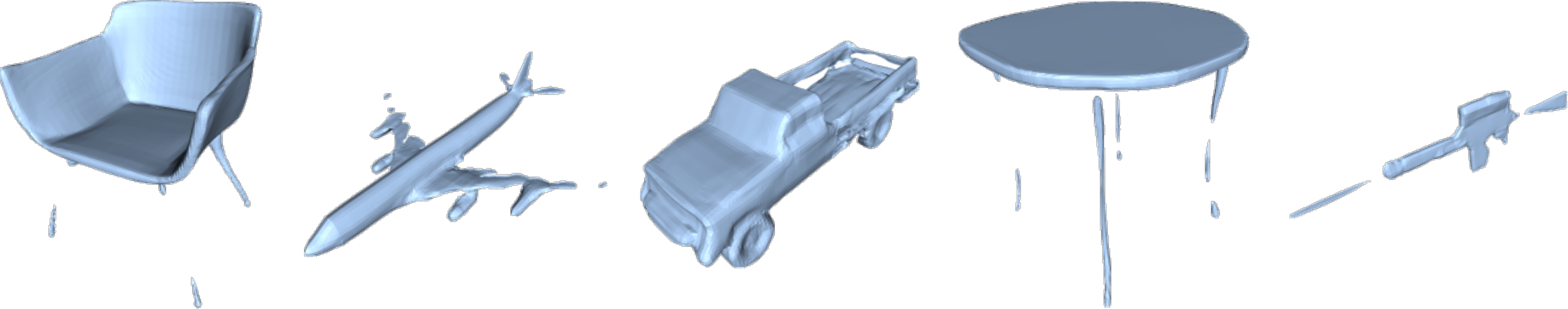}
    \end{overpic}
    \caption{Incomplete shapes generated by our method. The thin structures are not captured by the zero isosurfaces.}
    \label{fig:failure} \vspace{-4mm}
\end{figure}

\myparagraph{Limitations} A few limitations exist in our work. As there is no explicit shape part structure utilized in our framework design, we notice that a small portion of the generated shapes is not complete: tiny and thin parts could be missing, as shown in \cref{fig:failure}. Increasing the iso-value in extracting mesh surfaces can mitigate this issue, but there is no guarantee. We also found that complex geometry patterns such as various supporting beam layouts of chair backs, are not learned by our network, possibly due to unbalanced data distribution.

\myparagraph{Future work}
Some research directions are left for future exploration.
First, using truncated SDF would help leverage higher resolution ground-truth SDF during training to improve shape quality while reducing memory footprint and computational time. Second, our preliminary test on style mixing~\cite{karras2019style} reveals that there is some relation between the styles learned from the network and the semantic structures of the shapes, but it is still difficult to make a semantically meaningful disentanglement. Furthermore, it would be very useful to extend our model to 3D scene generation.

\bibliographystyle{eg-alpha-doi}
\bibliography{egbibsample}
\end{document}